\PassOptionsToPackage{table,xcdraw}{xcolor}
\documentclass[10pt,twocolumn,letterpaper]{article}

\usepackage[pagenumbers]{iccv} 

\definecolor{iccvblue}{rgb}{0.21,0.49,0.74}
\usepackage[pagebackref,breaklinks,colorlinks,allcolors=iccvblue]{hyperref}

\usepackage{graphicx}
\usepackage{amsmath}
\usepackage{amssymb}
\usepackage{booktabs}
\usepackage{comment}

\usepackage[table,xcdraw]{xcolor}
\usepackage{times}
\usepackage{epsfig}
\usepackage{float}
\usepackage{multirow}
\usepackage{amsfonts}
\usepackage{graphics}
\usepackage{witharrows}
\usepackage{arydshln}
\usepackage{amsfonts} 
\usepackage{mathtools}
\usepackage{enumitem}

\usepackage{algorithm}
\usepackage{algpseudocode}

\usepackage{subcaption}
\usepackage{comment}

\newcolumntype{g}{>{\columncolor[HTML]{EFEFEF}}c}
\newcolumntype{q}{>{\columncolor[HTML]{EFEFEF}}l}

\usepackage[capitalize]{cleveref}
\crefname{section}{Sec.}{Secs.}
\Crefname{section}{Section}{Sections}
\Crefname{table}{Table}{Tables}
\crefname{table}{Tab.}{Tabs.}

\newcommand{\fref}[1]{Figure \ref{#1}}
\newcommand{\tref}[1]{Table \ref{#1}}
\newcommand{\sref}[1]{Section \ref{#1}}

\def\paperID{10337} 
\def\confName{ICCV}
\def\confYear{2025}

\title{Visual {\color{red}Mod}ality {\color{red}Prompt} for Adapting Vision-Language Object Detectors}

\author{Heitor R. Medeiros$^*$\footnotetext{Specifically, I'd write comments in this one.}. \qquad Atif Belal \qquad Srikanth Muralidharan \\ Eric Granger \qquad Marco Pedersoli \\
LIVIA, Dept. of Systems Engineering, ETS Montreal, Canada\\
International Laboratory on Learning Systems (ILLS)}

\begin{document}

\twocolumn[{%
\renewcommand\twocolumn[1][]{#1}%
\maketitle
\begin{center}
\vspace{-.8cm}

    \centering
    \captionsetup{type=figure}
    \label{figure:initial_fig}
    \begin{subfigure}[!htp]{0.175\textwidth}
        \caption{GT}
    
        \makebox[0pt][r]{\makebox[18pt]{\raisebox{30pt}{\rotatebox[origin=c]{90}{LLVIP}}}}%
        \includegraphics[width=\textwidth]{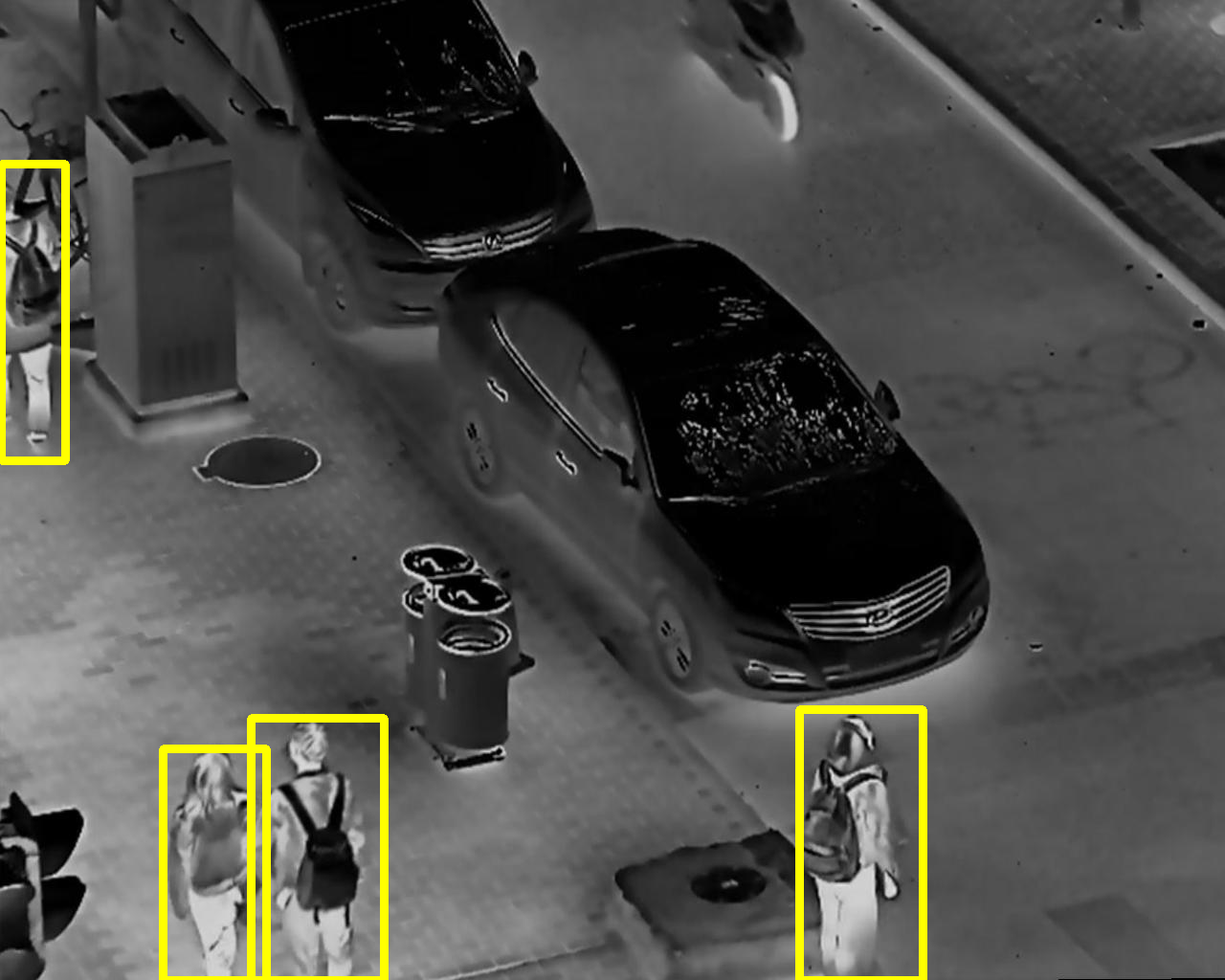}

        \makebox[0pt][r]{\makebox[18pt]{\raisebox{30pt}{\rotatebox[origin=c]{90}{FLIR}}}}%
        \includegraphics[width=\textwidth]{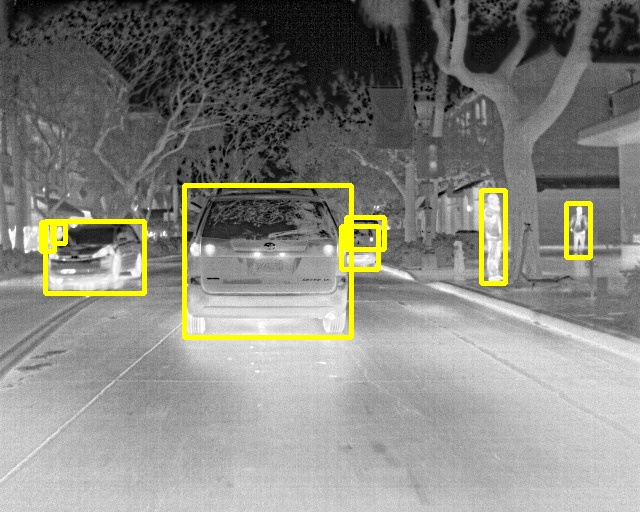}
        
        \makebox[0pt][r]{\makebox[18pt]{\raisebox{30pt}{\rotatebox[origin=c]{90}{NYU$_{v2}$}}}}%
        \includegraphics[width=\textwidth]{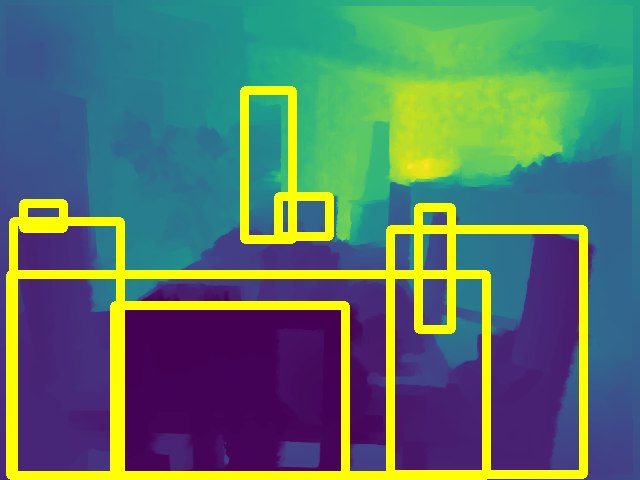}

    \end{subfigure}
    \begin{subfigure}[!htp]{0.175\textwidth}
        \caption{Zero-Shot}
    
        \includegraphics[width=\textwidth]{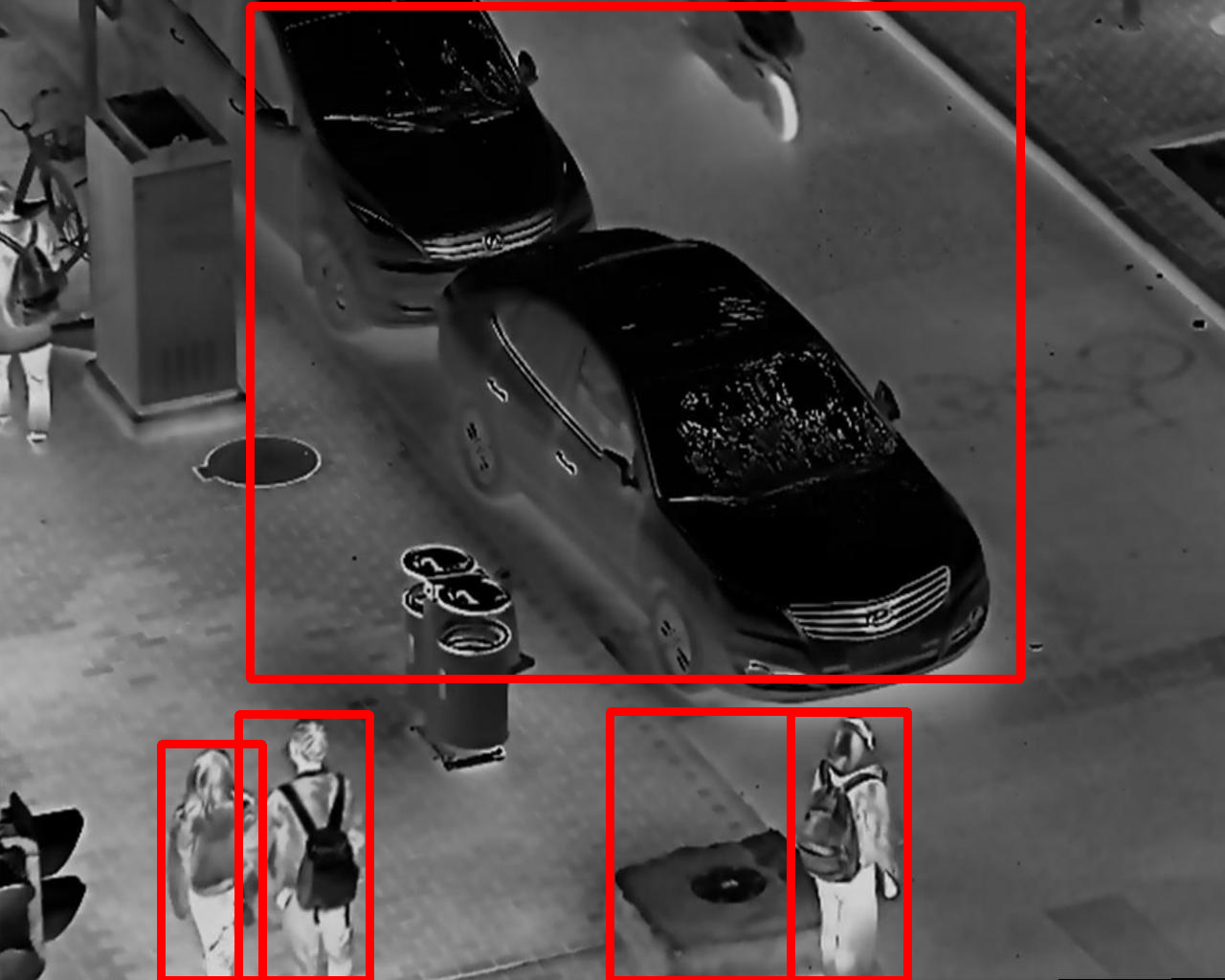}
    
        \includegraphics[width=\textwidth]{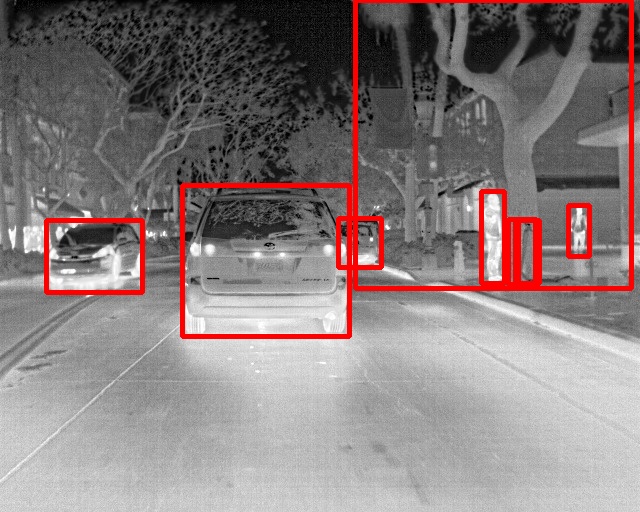}
        
        \includegraphics[width=\textwidth]{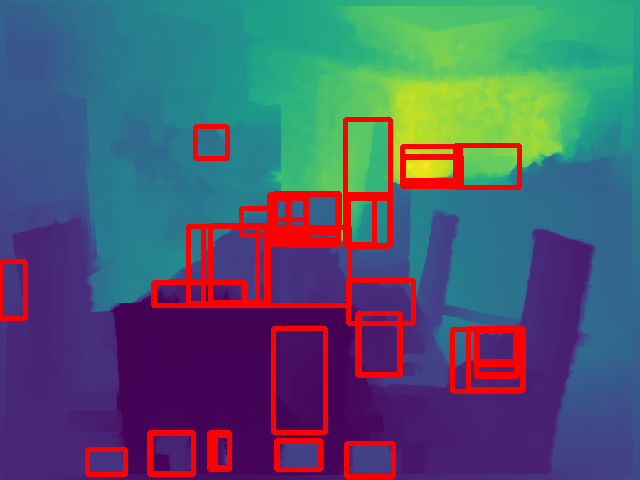}
        
    \end{subfigure}
    \begin{subfigure}[!htp]{0.175\textwidth}
        \caption{Visual Prompt}
        
        \includegraphics[width=\textwidth]{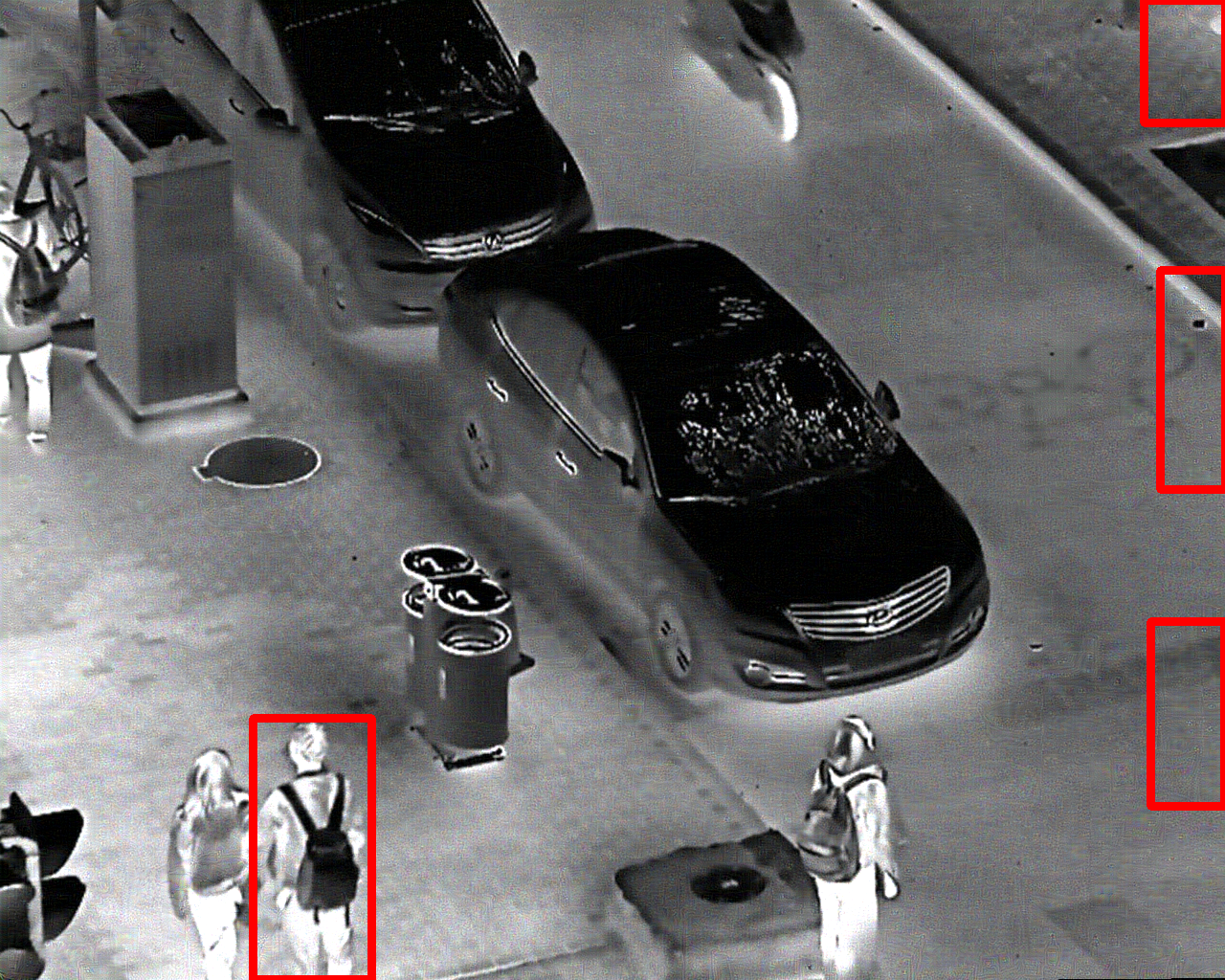}
        
        \includegraphics[width=\textwidth]{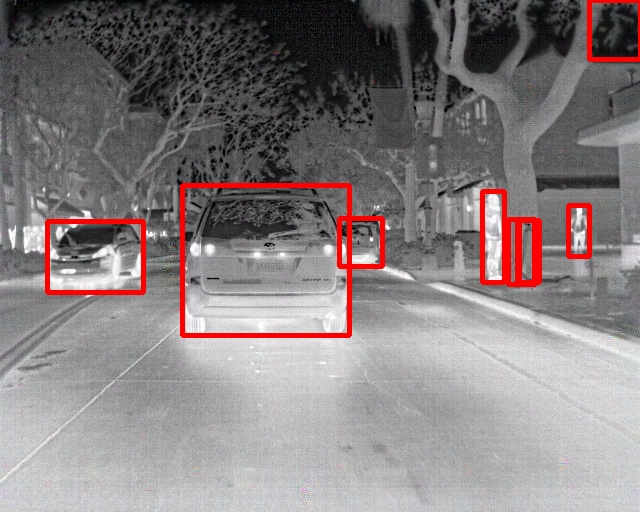}
        
        \includegraphics[width=\textwidth]{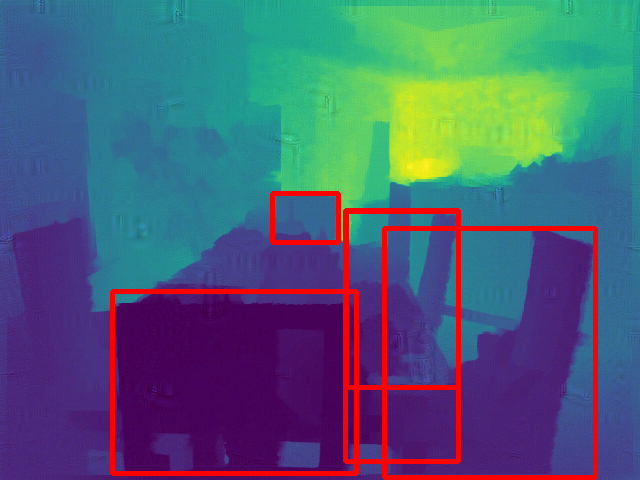}
        
    \end{subfigure}
    \begin{subfigure}[!htp]{0.175\textwidth}
        \caption{ModPrompt (Ours)}
    
        \includegraphics[width=\textwidth]{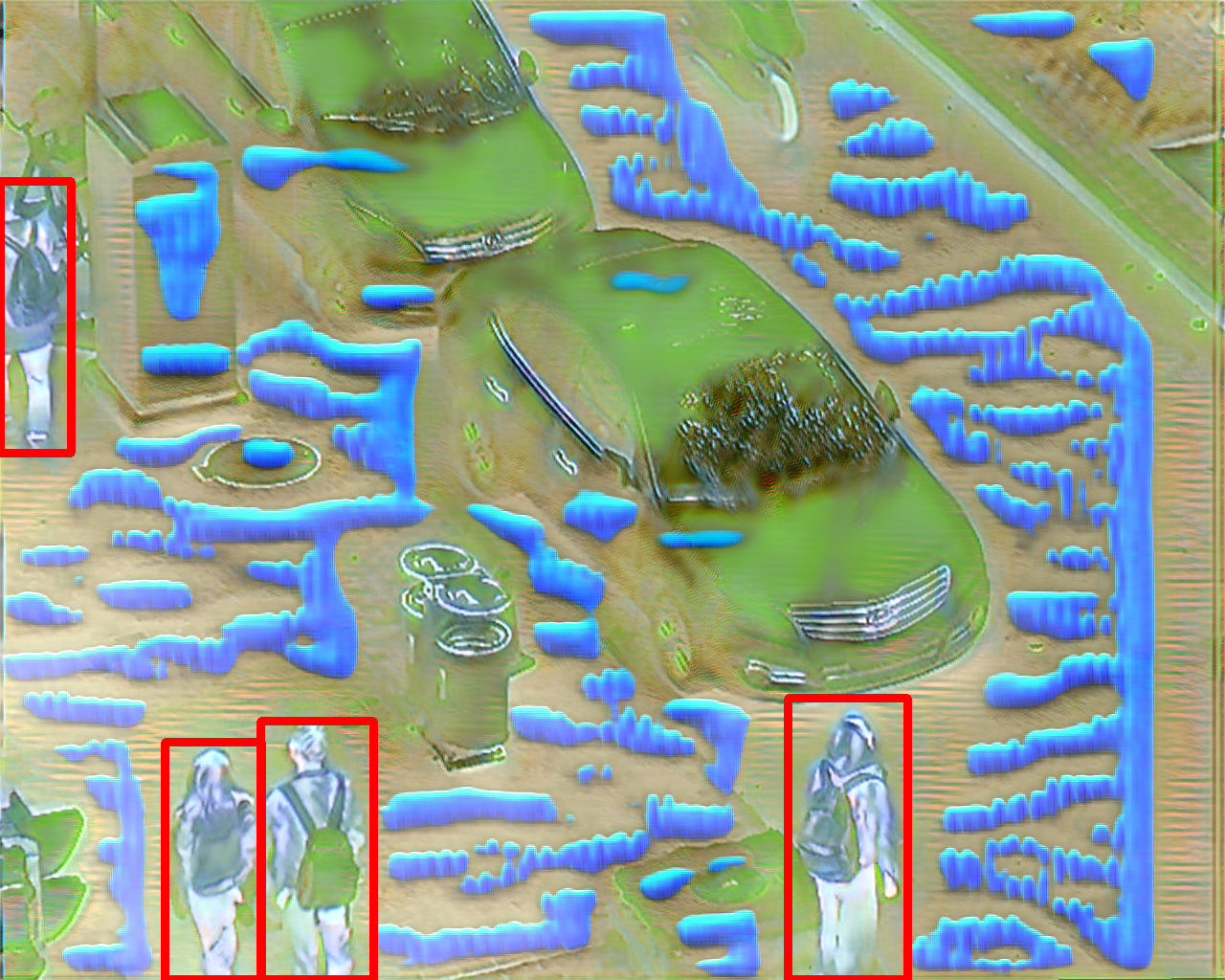}
        
        \includegraphics[width=\textwidth]{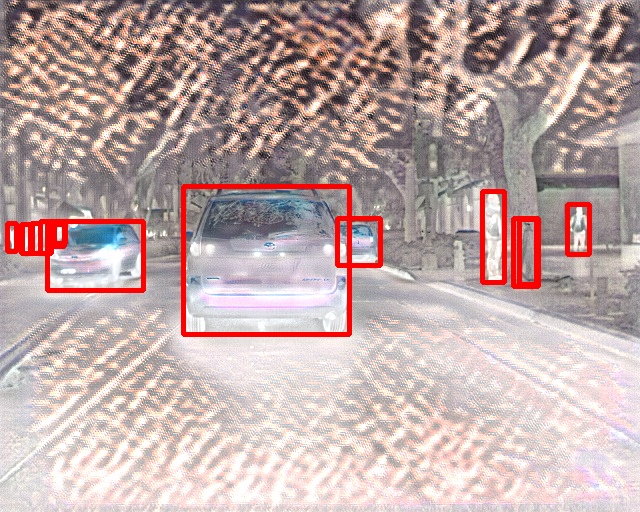}
        
        \includegraphics[width=\textwidth]{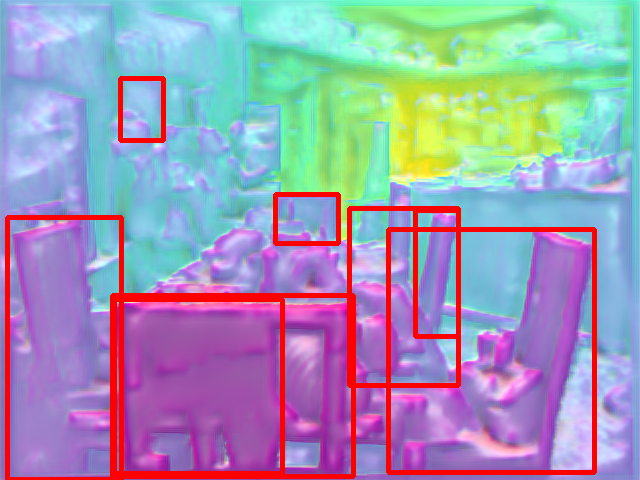}    
        
    \end{subfigure}

    \setcounter{figure}{0}
    \captionof{figure}{\textbf{Detections of different approaches across modalities}: LLVIP and FLIR datasets (infrared) and NYU$_{v2}$ (depth). Each column corresponds to a different approach: \textbf{(a) GT (Ground Truth):} Shows in yellow the ground-truth bounding boxes for objects. \textbf{(b) Zero-Shot:} Displays detections (in red) from a zero-shot model. This model has missed several detections and some inaccurate boxes without specific tuning. \textbf{(c) Visual Prompt:} Illustrates detections with a visual prompt added to the image. It shows improvements over zero-shot, with more accurate detection in certain areas, but still misses some objects. \textbf{(d) ModPrompt (Ours):} Detections from our proposed model. ModPrompt generates artifacts on the image to enhance objects and suppress background, facilitating detection.
}
\end{center}%
}]

\def\thefootnote{*}\footnotetext{Contact: heitor.rapela-medeiros.1@ens.etsmtl.ca}

\begin{abstract}
 The zero-shot performance of object detectors degrades when tested on different modalities, such as infrared and depth. While recent work has explored image translation techniques to adapt detectors to new modalities, these methods are limited to a single modality and apply only to traditional detectors. Recently, vision-language detectors, such as YOLO-World and Grounding DINO, have shown promising zero-shot capabilities, however, they have not yet been adapted for other visual modalities. Traditional fine-tuning approaches compromise the zero-shot capabilities of the detectors. The visual prompt strategies commonly used for classification with vision-language models apply the same linear prompt translation to each image, making them less effective. To address these limitations, we propose ModPrompt, a visual prompt strategy to adapt vision-language detectors to new modalities without degrading zero-shot performance. In particular, an encoder-decoder visual prompt strategy is proposed, further enhanced by the integration of inference-friendly modality prompt decoupled residual, facilitating a more robust adaptation. Empirical benchmarking results show our method for modality adaptation on two vision-language detectors, YOLO-World and Grounding DINO, and on challenging infrared (LLVIP, FLIR) and depth (NYU$_{v2}$) datasets, achieving performance comparable to full fine-tuning while preserving the model's zero-shot capability. Code available at: \url{https://github.com/heitorrapela/ModPrompt}.
\end{abstract}

\begin{figure*}[!ht]
\centering
    \resizebox{\textwidth}{!}{%

    \begin{tabular}{c}
    
    \includegraphics[width=1.0\textwidth]{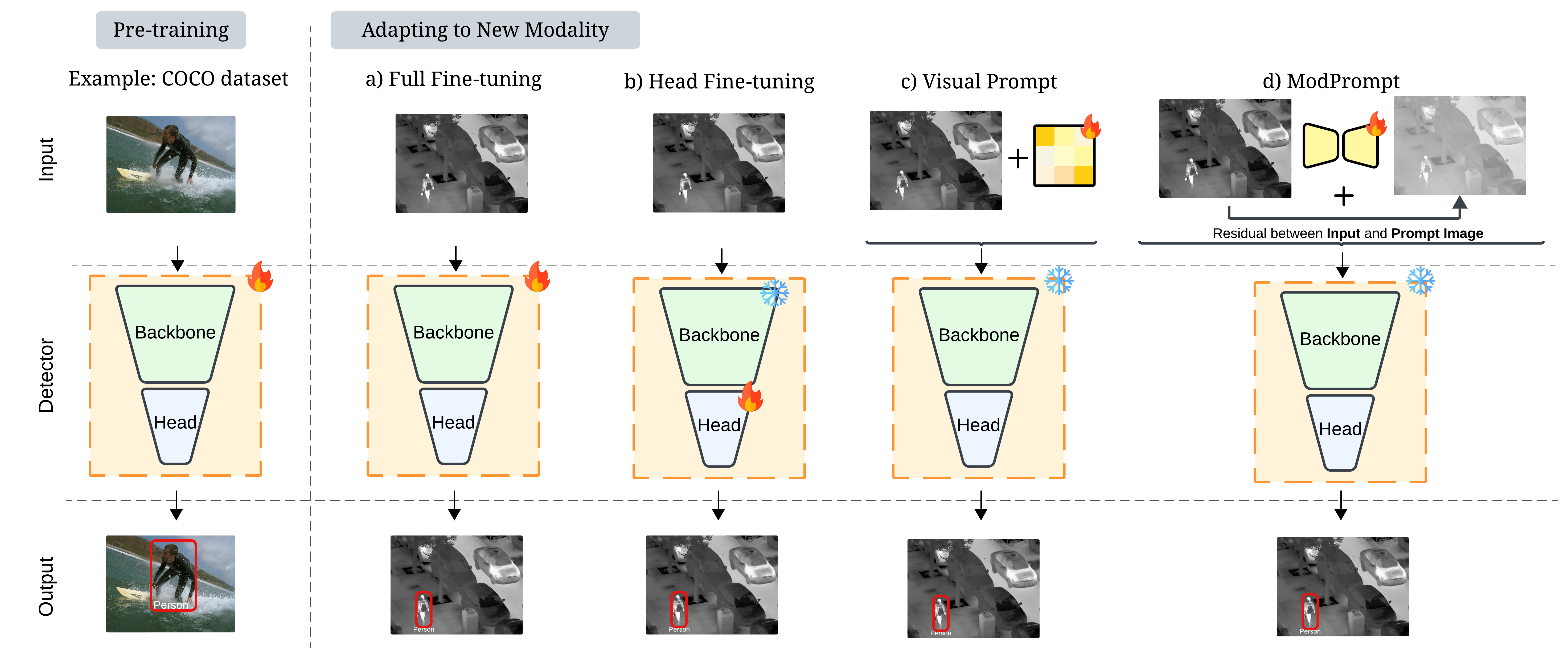} \\
    
    \end{tabular}
    }
\caption{Strategies to adapt object detectors to new modalities: \textbf{(a) Full Fine-tuning:} Both the backbone (the part of the model responsible for feature extraction) and the head (responsible for the final output, like object detection) are updated with new training data. \textbf{(b) Head Fine-tuning:} Only the head is fine-tuned while the backbone remains frozen. \textbf{(c) Visual Prompt:} Uses a visual prompt added to the input. The backbone and head remain unchanged, but the visual prompt guides the model to better interpret the new modality. \textbf{(d) Our Modality Prompt.} Similarly to a visual prompt, the input image is added to a visual prompt. The main difference is that here the prompt is not static, it is a transformation of the input image.}
\label{fig:different_modality_adaptation}
\vspace{-.4cm}
\end{figure*}

\section{Introduction}
\label{sec:intro}

Object detection (OD) is a key challenge in computer vision, aiming to localize and classify objects in images~\citep{zou2023object}. Recent OD advancements, driven by applications like autonomous driving~\citep{stilgoe2018machine, michaelis2019benchmarking}, surveillance~\citep{ramachandran2021review, dubail2022privacy}, and robotics~\citep{eitel2015multimodal, pierson2017deep, ivorra2018multimodal}, have improved real-world systems significantly. Although traditional detectors like YOLO~\citep{yolov8_ultralytics} or DINO~\citep{zhang2022dino} perform well, they are limited to fixed vocabularies, such as the $80$ categories in the COCO~\citep{lin2014microsoft} dataset, and exhibit poor zero-shot performance~\citep{bansal2018zero}. These detectors are trained on predefined categories, restricting their adaptability to varying scenarios. In contrast, Vision-Language Models (VLMs) combine visual representations and semantic text embeddings, allowing them to dynamically understand pixel-level features in context~\cite{radford2021learning}. VLMs enhance accuracy and flexibility by recognizing unseen objects through the integration of language, paving the way for open-vocabulary detection~\citep{zang2022open}, such as Grounding DINO~\citep{liu2023grounding} and YOLO-World~\citep{cheng2024yolo}.

Generally, these open-vocabulary detectors are pre-trained on large-scale RGB datasets, making them capable of zero-shot detection~\cite{zhang2022glipv2, liu2023grounding}. However, their performance degrades if the domain shift is large, as in a different modality, such as infrared or depth. A common approach to mitigate this is by finetuning the VLMs on the downstream modality dataset. However, it makes the VLMs lose their zero-shot capabilities due to catastrophic forgetting~\citep{efficientprompttuning}. Some techniques have been explored to adapt the VLMs to downstream tasks for image classification. These include text-prompt tuning (TPT)~\citep{zhou2022learning, zhou2022conditional} and visual prompt tuning (VPT)~\citep{jia2022visual, bahng2022exploring}. Unlike traditional finetuning, prompt tuning involves adding learnable prompts to the model's input, allowing it to remain unchanged, and preserve its zero-shot capability. VPT methods for classification learn visual prompt tokens to adapt the classifier of transformer-based models to the downstream tasks. Visual prompts are effective when adapting VLMs to a downstream task of the same modality (normally RGB). However, these methods are also less effective if the downstream data is from another visual modality with a large modality shift between pre-trained and downstream data, as shown in~\tref{tab:main_yolo_world}.

In addition to incorporating language into detectors, another major driver of advancements in real-world applications is the integration of diverse visual modalities, such as infrared~\citep{medeiros2024mipa, wang2022improving, li2021yolo} and depth~\citep{li2023bevdepth, xu2017multi, yang2022deepinteraction}. These modalities provide additional spatial information, enhancing visibility in low-light and obstructed conditions, resulting in more robust and accurate object detectors~\citep{bustos2023systematic}. Adapting detectors for different modalities unlocks their potential in specialized contexts by combining the efficiency of transfer learning with modality-specific enhancements. This approach bridges the gap between general-purpose detectors capable of zero-shot detection and the nuanced demands of novel modalities, where labels are still scarce compared to RGB data.

Therefore, to overcome limitations in modality adaptation, image translation~\citep{herrmann2018cnn, ozkanouglu2022infragan} methods have been explored for traditional detector-based models. However, these approaches primarily optimize the image translation loss rather than directly focusing on improving detection performance. More recent approaches, such as HalluciDet~\citep{medeiros2024hallucidet} and ModTr~\citep{medeiros2024modality}, have demonstrated improvement by emphasizing detection loss optimization during adaptation. However, these methods are limited to traditional detectors without language mechanisms, and they did not explore more challenging modalities such as depth. For HalluciDet, the prior pre-trained knowledge is lost during the adaptation phase. In ModTr, the COCO pre-trained knowledge is preserved but it lacks the zero-shot powerful capability of VLMs. In contrast, our work leverages recent vision-language detectors, incorporating textual information and allowing for adaptation to different modalities, exploring the benefits of both visual and text information for improving modality adaptation.

In this paper, we propose ModPrompt, a visual prompt strategy to adapt vision-language ODs in the input space to new visual modalities without sacrificing zero-shot knowledge of the vision-language detector. ModPrompt is a detector-agnostic encoder-decoder visual prompt strategy that can easily be integrated with any VLM detector, regardless of the detector's backbone type. We also introduce an inference-friendly residual mechanism named Modality Prompt Decoupled Residual (MPDR) capable of adapting the learnable text embeddings with the ModPrompt loss, while preserving full knowledge of the model due to the decoupled embedding parameters. We demonstrate that the combination of MPDR with ModPrompt, improves the performance of the detector across different modalities while retaining the original zero-shot knowledge, both in the vision and language components. To the best of our knowledge, this is the first work that focuses on adapting state-of-the-art VLM-based OD models to new modalities. Furthermore, ModPrompt is validated on two state-of-the-art VLM ODs, Grounding DINO and YOLO-World, and across two widely used visual modalities, infrared and depth.

\noindent \textbf{Our main contributions can be summarized as follows.} 

\noindent \textbf{(1)} We introduce ModPrompt, a novel pixel-level prompt adaptation strategy based on input translation. Our method is backbone-agnostic, leading to improved adaptability of vision-language detectors across different modalities and preserves the vision encoder knowledge. Additionally, we introduce MPDR, a text-embedding modality prompt decoupled residual mechanism, which preserves full knowledge of the text-embedding while increasing the performance of the model for the novel modality.

\noindent \textbf{(2)} We conduct a detailed investigation of various pixel-level visual prompt techniques, offering insights into why traditional prompt methods struggle with adapting object detectors to new modalities. Our analysis demonstrates how ModPrompt overcomes these limitations to significantly enhance detection performance.

\noindent \textbf{(3)} We provide a comprehensive benchmark on the visual prompts for modality adaptation across two primary open-vocabulary OD models, YOLO-World and Grounding DINO. We evaluate our method across infrared and depth datasets, achieving in some cases a level of performance comparable to full fine-tuning while preserving the model's zero-shot capability.

\setlength{\tabcolsep}{8pt}
\renewcommand{\arraystretch}{1.0}


\begin{table*}[!ht]
    \centering
    \resizebox{1.0\textwidth}{!}{%
    \begin{tabular}{lqgggggg} 
        \toprule
        \rowcolor{white}
        \multirow{2}{*}[-0.7em]{\textbf{Dataset}} & \multirow{2}{*}[-0.7em]{\textbf{Method}} &  \multicolumn{3}{c}{\textbf{YOLO-World}} &  \multicolumn{3}{c}{\textbf{Grounding DINO}} \\
        \cmidrule(lr){3-5}
        \cmidrule(lr){6-8}
        \addlinespace[5pt]   

        \rowcolor{white}
        {} & {} & \multicolumn{1}{c}{\multirow{2}{*}[1em]{\textbf{AP$_{50}$}}} & 
        \multicolumn{1}{c}{\multirow{2}{*}[1em]{\textbf{AP$_{75}$}}} &
        \multicolumn{1}{c}{\multirow{2}{*}[1em]{\textbf{AP}}} & \multicolumn{1}{c}{\multirow{2}{*}[1em]{\textbf{AP$_{50}$}}} & 
        \multicolumn{1}{c}{\multirow{2}{*}[1em]{\textbf{AP$_{75}$}}} &
        \multicolumn{1}{c}{\multirow{2}{*}[1em]{\textbf{AP}}}   \\
        \midrule

        {} & Zero-Shot (ZS)  & 81.00 ± 0.00 & 57.80 ± 0.00 & 53.20 ± 0.00 & 85.50 ± 0.00 & 62.70 ± 0.00 & 56.50 ± 0.00 \\

        \rowcolor{white}
        {} & Head Finetuning (HFT) & 93.57 ± 0.05 &	73.83 ± 0.19 &	64.80 ± 0.08 & 87.53 ± 0.06 & 65.57 ± 0.23 & 58.10 ± 0.20  \\
        
        {} & Full Finetuning (FT) & 97.43 ± 0.05 & 77.93 ± 0.21 & 67.73 ± 0.09  & 97.17 ± 0.31 & 79.93 ± 0.83 & 67.83 ± 0.96 \\
        
        \cmidrule(lr){2-8}
        \rowcolor{white}
        {} & Visual Prompt (Fixed) & 70.30 ± 7.89 &  45.67 ± 6.97 &  43.53 ± 5.79 & 83.83 ± 0.06 &  61.53 ± 0.23 &  55.13 ± 0.15 \\  
        
        \multirow{4}{*}[2em]{\textbf{LLVIP - IR}} & Visual Prompt (Random) & 60.13 ± 0.29 & 38.73 ± 0.17 &  36.87 ± 0.12  & 83.87 ± 0.06 & 61.37 ± 0.06 &  55.03 ± 0.06 \\  
        \rowcolor{white}
        {} & Visual Prompt (Padding) & 79.87 ± 1.00 &    51.77 ± 0.90 &  49.30 ± 0.83 & 82.73 ± 0.31 &    60.00 ± 0.35 &  55.13 ± 0.15 \\


        {} & Visual Prompt (WM) & 82.00 ± 1.59 & 53.90 ± 1.06 &  50.90 ± 0.94  & 69.57 ± 0.93 & 41.37 ± 1.27 &  40.77 ± 0.87 \\
        
        \rowcolor{white}
        {} & Visual Prompt (WM$_{v2}$) & 74.10 ± 0.43 & 46.47 ± 0.62 &   44.70 ± 0.22 & 69.87 ± 1.12 &  41.77 ± 1.30 &  41.13 ± 0.96 \\

        \cmidrule(lr){2-8}
        {} & \textbf{ModPrompt (Ours)} & \textbf{92.80 ± 0.29} & \textbf{70.73 ± 1.02} & \textbf{62.87 ± 0.63}  & \textbf{93.13 ± 0.15} & \textbf{67.17 ± 0.78} & \textbf{60.10 ± 0.50} \\
        
        \midrule
        \addlinespace[2pt]        
        \midrule
        
        {} & Zero-Shot (ZS)  & 04.80 ± 0.00 &    03.10 ± 0.00 &  03.00 ± 0.00  & 08.30 ± 0.00 &    05.60 ± 0.00 &  05.30 ± 0.00 \\

        \rowcolor{white}
        {} & Head Finetuning (HFT) & 21.03 ± 0.12 & 12.03 ± 0.37 &	12.00 ± 0.14 & 18.24 ± 0.85 & 13.23 ± 0.46 & 12.10 ± 0.21 \\

        {} & Full Finetuning (FT) & 49.90 ± 0.08 &  36.40 ± 0.29 &  33.57 ± 0.26  & 51.60 ± 2.09 &  39.17 ± 1.88 &  35.77 ± 1.70 \\
        
        \cmidrule(lr){2-8}  
        \rowcolor{white}
        {} & Visual Prompt (Fixed) &  04.67 ± 0.05 &    03.07 ± 0.05 &  02.90 ± 0.00  & 08.27 ± 0.06 &    05.57 ± 0.06 &  05.27 ± 0.06 \\

        \multirow{4}{*}[2em]{\textbf{NYU$_{v2}$ - Depth}} & Visual Prompt (Random) & 04.23 ± 0.12 &    02.63 ± 0.05 &  02.53 ± 0.05  & 08.33 ± 0.06 & 05.53 ± 0.06 &  05.27 ± 0.06 \\
        
        \rowcolor{white}
        {} & Visual Prompt (Padding) &  03.97 ± 0.05 &  02.50 ± 0.00 &  02.43 ± 0.05  &  07.63 ± 0.06 &  05.17 ± 0.06 &  04.63 ± 0.06 \\


        {} & Visual Prompt (WM) & 10.67 ± 0.12 &    06.90 ± 0.22 &  06.57 ± 0.05   & 04.87 ± 0.06 &    02.97 ± 0.06 &  02.97 ± 0.06 \\
        
        \rowcolor{white}
        {} & Visual Prompt (WM$_{v2}$) &   10.63 ± 0.12 &  06.67 ± 0.12 &  06.50 ± 0.08  &   05.00 ± 0.10 &  03.07 ± 0.06 &  03.03 ± 0.06 \\

        \cmidrule(lr){2-8}
        {} & \textbf{ModPrompt (Ours)} & \textbf{37.17 ± 0.57} & \textbf{27.50 ± 0.64} & \textbf{24.93 ± 0.50}  &  \textbf{21.70 ± 0.20} & \textbf{15.03 ± 0.29} & \textbf{14.13 ± 0.21} \\

        \bottomrule
    \end{tabular}
    }
    \caption{Detection performance (APs) for YOLO-World and Grounding DINO for the two main datasets evaluated: LLVIP-IR and NYU$_{v2}$-Depth. The different visual prompt adaptation techniques are compared with our ModPrompt, and the zero-shot (ZS), head finetuning (HFT), and full finetuning (FT) are also reported, where the full finetuning is the upper bound.}
    \label{tab:main_yolo_world}
    \vspace{-.4cm}
\end{table*}

\section{Related Work}
\label{sec:related_work}

\noindent \textbf{Open-Vocabulary Object Detection.} Recent advancements in open-vocabulary OD leverage vision-language models (VLMs) and multimodal alignment at the region level to enable novel concept recognition. For instance, RegionCLIP~\citep{zhong2022regionclip} introduces region-level pseudo labels for contrastive pretraining, effectively supporting both open-vocabulary and zero-shot OD. ~\citet{gu2021open} frame open-vocabulary detection as knowledge distillation, aligning with CLIP’s teacher of visual-text region embeddings. ~\citet{lin2022learning} use bipartite matching for image-text alignment and validate their model on open-vocabulary benchmarks. GLIP~\citep{li2022grounded} employs deep multimodal fusion with self-generated box data, demonstrating robust zero-shot detection. GLIPv2~\citep{zhang2022glipv2} further enhances vision language tasks using region-grounded pretraining and shows improvements by contrastive learning with other image regions. More recently, Grounding DINO~\cite{liu2023grounding} pretrains a multiphase image-text fusion module for both open-vocabulary and referring OD and YOLO-World~\cite{cheng2024yolo} introduces a reparametrizable fusion approach that enables efficient inference in a lightweight model. 

\noindent \textbf{Downstream Adaptation of Large Vision Models.} Recent seminal works explore ways to adapt large-scale vision models to downstream tasks without losing prior knowledge and perform adaptation in parameter-efficient ways. Co-op~\cite{zhou2022learning} learns a continuous representation of context prompt in the text space for downstream task learning. Visual prompt Tuning~\citep{jia2022visual} explores adding training parameters at different stages in the network, demonstrating competitive performance across multiple recognition tasks. Bahng et al.~\citep{bahng2022exploring} adapt large-scale vision models to new tasks by performing linear probing at the image space once. CLIP-Adapter~\cite{gao2024clip} adapt in feature space for downstream tasks through bottleneck layer and residual blending instead of adapting at the input side. Yu et al.~\cite{yu2023task} improve downstream task performance while decoupling the new parameters from the model using task residual learning for classification tasks. 

\noindent \textbf{Image Translation for Detection.} Image translation methods map images from one domain to another by adjusting the domain characteristics while preserving essential content~\cite{pang2021imagetoimage}. Most methods rely on generative approaches~\cite{kingma2022autoencoding, goodfellow2014generative, pang2021imagetoimage}. A notable method, Pix2Pix by Isola et al., uses a paired generator-discriminator setup to generate domain-specific images based on paired data and labeled guidance~\cite{isola2017image}. On the other hand, CycleGAN by Zhu et al. introduced a framework for domain translation without paired supervision, making it effective for tasks with unaligned datasets~\cite{CycleGAN2017}. For OD, HalluciDet~\cite{medeiros2024hallucidet} adapt image translation methods in a two-step adaptation; in the first, the detector is fine-tuned in an RGB distribution closer to the target IR modality, subsequently, the RGB detector is used to train the translator model. Recently, ModTr~\cite{medeiros2024modality} proposed to train the translator using zero-shot detectors pre-trained on COCO combined with fusion strategies, which benefits the IR adaptation while preserving the pre-trained COCO knowledge. Different from these methods, we focus on multimodality adaptation from large zero-shot vision-language ODs to other modalities such as infrared and depth, leveraging the powerful generalization abilities of these RGB-based open-vocabulary detectors without losing their prior knowledge.


\section{Proposed Method} 
\label{sec:method}

\subsection{Preliminary Definitions}

\noindent \textbf{Vision Language Object Detection.} The training dataset for traditional object detectors can be represented as $\mathcal{D}=\{(X, Y)\}_{j=1}^{M}$. Here, $X\in\mathbb{R}^{W\times H\times C}$ represents an image with dimensions $W\times H$ and $C$ channels, and $Y=\{(b_i, c_i)\}_{i=1}^{N}$ consist of bounding boxes $b_i$ and object category $c_i$. For the visual language model, the annotations are reformulated as region-text pair $Y=\{(b_i, t_i)\}_{i=1}^{N}$, where $t_i$ corresponds to the text for the region $b_i$. Specifically, $t_i$ is the name of the object category in $b_i$. The objective of the object detection task is to accurately identify all objects in a given image. The average precision (AP) metric across classes is used as the standard evaluation protocol. An object detector is formally represented as the mapping $f_{\theta}: \mathbb{R}^{W \times H \times C} \to \hat{Y}$, where $\theta$ denotes the parameter vector. The detection cost function $\mathcal{C}_{det}(\theta)$, a differentiable proxy for the AP metric, is computed as an average loss of the detection loss $\mathcal{L}_{det}$ over the entire dataset $\mathcal{D}$, described mathematically as:
\begin{equation}
    \begin{split}
        \mathcal{C}_{det}(\theta)=\frac{1}{|\mathcal{D}|} \sum_{(x, Y) \in \mathcal{D}} \mathcal{L}_{det}(f_\theta(x), Y).
    \end{split} 
    \label{eq:detection_loss}
\end{equation}

\noindent \textbf{Visual Prompt for OD.} Given a frozen pre-trained detector, the objective of Visual Prompt method is to learn a task-specific visual prompt $v_{\phi}$, parametrized by $\phi$, which can be combined with the input $x$ to improve the final detection performance. During test time, this visual prompt is added to the test images to adapt it for the desired detection task. As described in the following equation:
\begin{equation}
    \begin{split}
        \mathcal{C}_{\text{vp}}(\phi)=\frac{1}{|\mathcal{D}|} \sum_{(x, Y) \in \mathcal{D}} \mathcal{L}_{det}(f_\theta(x+v_{\phi}), Y).
    \end{split} 
\label{eq:visual_prompt_od_cost}
\end{equation}

\subsection{ModPrompt} The visual prompt methods discussed above learn simple linear transformation, like adding a fixed patch to each image. These transformations are learned during the training process and are not conditioned on the input image at inference, making them less effective. In ModPrompt, we incorporate a function $h_{\vartheta}$, an encoder-decoder inspired by U-Net~\citep{ronneberger2015u}, dependent on the input image $x$, which is trained conditioned on labels $Y$. Our encoder-decoder is flexible in terms of the encoder backbone initialized from pre-trained weights. We adapt the last layer of the decoder to constrain it to be $3$-channels with values ranging between $0$ and $1$ to simulate a pseudo-RGB image. Note, that this mechanism is completely different from traditional U-Nets, which are trained to reconstruct and generate $c$-channels for semantic segmentation tasks, requiring both input and ground truth segmentation map during training. However, ModPrompt requires access only to the pre-trained RGB detector and final target modality data, so it is based on guidance detection loss instead of reconstruction. The ModPrompt training cost is defined by the following equation:

\begin{equation} 
\begin{split}
        \mathcal{C}_{\text{mp}}(\vartheta)=\frac{1}{|\mathcal{D}|} \sum_{(x, Y) \in \mathcal{D}} \mathcal{L}_{det}(f_\theta(x + h_\vartheta(x)), Y). 
\end{split}
\label{eq:modprompt}
\end{equation}

This translation function ($f_\theta$) is responsible for adapting the input image to a modality representation optimally suited to address detection in the target modality. Our idea of a visual ModPrompt is inspired by text conditional prompt~\citep{zhou2022conditional} work, which incorporates a small network for the text encoder conditional adaptation. In our approach, the visual prompt depends on the input image.

\begin{figure}[h]
\centering
    \resizebox{\columnwidth}{!}{%

    \begin{tabular}{c}
    
    \includegraphics[width=1.0\textwidth]{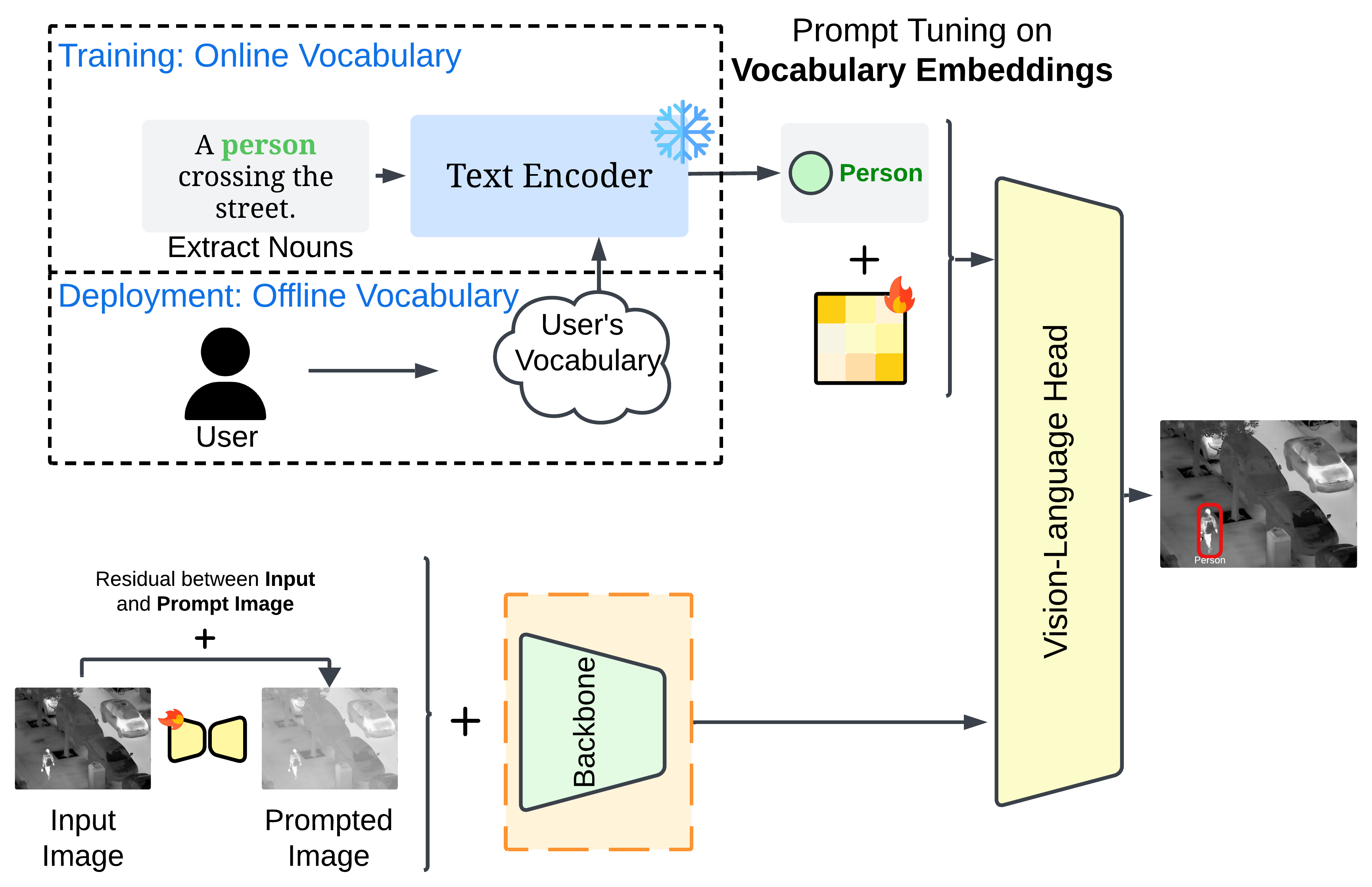} \\
    
    \end{tabular}
    }
\caption{Our proposed strategy for text-prompt tuning: \textbf{an inference-friendly} and \textbf{knowledge-preserving decoupled embedding tuning} method. An offline embedding is generated for each object category, and then a novel decoupled residual trainable parameters and the ModPrompt are integrated into the detector to adapt it to new modalities.}
\label{fig:text_embedding_tuning}
\vspace{-.6cm}
\end{figure}

\subsection{Modality Prompt Decoupled Residual with Knowledge Preservation (MPDR)} The vision-language models take both text and image as input. ModPrompt aims to transform the input image to adapt to the new modality. For the adaptation of the textual part, text prompt tuning methods~\citep{zhou2022learning, zhou2022conditional} have been explored for classification. But, these methods require back-propagating through the whole text backbone making them computationally expensive. Recently, task residual~\cite{yu2023task} was proposed that tunes the precomputed text embedding of the vision-language classifiers. Inspired by this, we design a strategy to efficiently tune the embedding of the vision-language detectors. As illustrated in~\fref{fig:text_embedding_tuning}, offline text embeddings are generated for each object category. Then, the MPDR, which are learnable embedding tokens, are tuned to adapt to a specific modality end-to-end with ModPrompt loss. The MPDR is crucial for decoupling the knowledge from the original embedding, which preserves prior text-embedding knowledge and is able to incorporate new knowledge during the adaptation stage. During the training, the loss of ModPrompt and MPDR work in a synergy way to adapt to any new modality. In the test phase, the MPDR can be deactivated easily by zero-masking its parameters, which is the same as full zero-shot embedding knowledge, or enabled which is the adapted embedded for this modality (that's why decoupling the knowledge is important) and doesn't introduce any inference overhead. The following equation describes the final training cost:

\begin{equation} 
\begin{split}
        \mathcal{C}_{\text{mp-tp}}(\vartheta, \phi)= \mathcal{C}_{\text{mp}}(\vartheta) + \mathcal{C}_{\text{tp}}(\phi),
\end{split}
\label{eq:modprompt_embeddingtuning}
\end{equation}
\noindent where $\mathcal{C}_{\text{mp}}(\vartheta)$ is the optimization of parameters $\vartheta$ of the encoder-decoder function, and $\mathcal{C}_{\text{tp}}(\phi)$ is the text-prompt adaptation on the embedding space by training new parameters $\phi$ for the class embeddings in the offline vocabulary.

\subsection{Training Summary}

For training the ModPrompt with MPDR, we first pre-compute the text embeddings for the target classes, using the text encoder. Then, we sum the learnable embedding residuals with the frozen embeddings and train it together with the ModPrompt. Note that different from other embedding adaptation approaches, our MPDR can be used with ModPrompt for full zero-shot knowledge preservation due to the residual decoupled embedding learning strategy. The cost function is based on Equation~\ref{eq:modprompt_embeddingtuning}, but instead of the online cost function for text $\mathcal{C}_{\text{tp}}(\phi)$, we work on the text embedding space, which is better suited for efficient training and inference.




\section{Results and Discussion}  
\label{sec:experiments}

\begin{table*}[t]
    \centering
    \resizebox{1.0\textwidth}{!}{%
    \begin{tabular}{lqgggqgg}
        \toprule
        \rowcolor{white}
        {\multirow{2}{*}[-0.7em]{\textbf{Detector}}} & \multirow{2}{*}[-0.7em]{\textbf{Method}}  &  \multicolumn{3}{c}{\textbf{LLVIP-IR}} &   \multicolumn{3}{c}{\textbf{NYU$_{v2}$-DEPTH}} \\

        \cmidrule(lr){3-5}
        \cmidrule(lr){6-8}
        \addlinespace[5pt]        
        \rowcolor{white}
        {}  & {} & \multicolumn{1}{c}{\multirow{2}{*}[1em]{\textbf{AP$_{50}$}}} & 
        \multicolumn{1}{c}{\multirow{2}{*}[1em]{\textbf{AP$_{75}$}}} &
        \multicolumn{1}{c}{\multirow{2}{*}[1em]{\textbf{AP}}} & \multicolumn{1}{c}{\multirow{2}{*}[1em]{\textbf{AP$_{50}$}}} & 
        \multicolumn{1}{c}{\multirow{2}{*}[1em]{\textbf{AP$_{75}$}}} &
        \multicolumn{1}{c}{\multirow{2}{*}[1em]{\textbf{AP}}}   \\
        \midrule

        {} &  \textbf{Fixed} & 86.60 ± 0.00 (+16.3) &   63.53 ± 0.05 (+17.8) &  57.67 ± 0.05 (+14.1) &  12.47 ± 0.05 (+7.80) &    07.50 ± 0.00 (+4.43) &   07.37 ± 0.05 (+4.47) \\

        \cmidrule(lr){2-8}
        \rowcolor{white}
        \multirow{2}{*}[-1.2em]{\textbf{YOLO-World}} & \textbf{Random} & 86.43 ± 0.05 (+26.3) & 63.40 ± 0.08 (+24.6) & 57.50 ± 0.00 (+20.6) & 12.23 ± 0.05 (+8.00) &  07.30 ± 0.08 (+4.67) &  07.20 ± 0.00 (+4.67) \\

        \cmidrule(lr){2-8}
        {} & \textbf{Padding} & 83.57 ± 0.12 (+3.70) &  59.03 ± 0.05 (+7.26) &  54.23 ± 0.05 (+4.93) &  10.63 ± 0.09 (+6.66) &  05.60 ± 0.08 (+3.10) &  05.97 ± 0.05 (+3.54) \\

        \cmidrule(lr){2-8}
        \rowcolor{white}
        {} & \textbf{WeightMap}  & 87.47 ± 0.17 (+5.47) &   62.23 ± 0.31 (+8.33) &  57.43 ± 0.05 (+6.53) & 13.43 ± 0.09 (+2.76) &  08.07 ± 0.12 (+1.17) &  07.90 ± 0.08 (+1.33) \\

        \cmidrule(lr){2-8}
        {} & \textbf{ModPrompt} & \textbf{96.60 ± 0.37 (+3.80)} & \textbf{77.27 ± 0.33 (+6.54)} &   \textbf{67.37 ± 0.05 (+4.50)} & \textbf{44.67 ± 0.17 (+7.50)} & \textbf{32.53 ± 0.66 (+5.03)} & \textbf{29.93 ± 0.12 (+5.00)} \\

        \midrule
        \addlinespace[2pt]        
        \midrule

        {} & \textbf{Fixed}  & 88.70 ± 0.66 (+4.87) & 67.73 ± 0.50 (+6.20)  & 60.13 ± 0.45 (+5.00) & 10.00 ± 0.00 (+1.73) & 06.40 ± 0.28 (+0.83) & 06.40 ± 0.28 (+1.13) \\

        \cmidrule(lr){2-8}
        \rowcolor{white}
        \multirow{2}{*}[-1.2em]{\textbf{Grounding DINO}} & \textbf{Random}  & 88.80 ± 0.95 (+4.93) & 68.00 ± 1.21 (+6.63)  & 60.37 ± 0.90 (+5.33) & 09.90 ± 0.14 (+1.57) & 06.50 ± 0.00 (+0.97) & 06.15 ± 0.07 (+0.88) \\
        
        \cmidrule(lr){2-8}
        {} & \textbf{Padding}  & 86.93 ± 0.55 (+4.20) & 65.23 ± 1.24 (+5.23)  & 58.07 ± 1.10 (+4.17) & 09.27 ± 0.15 (+1.64) & 06.00 ± 0.17 (+0.83) & 05.57 ± 0.15 (+0.87) \\

       \cmidrule(lr){2-8}
       \rowcolor{white}
        {} & \textbf{WeightMap} & 74.90 ± 0.92 (+5.33) & 46.80 ± 0.70 (+5.43)  & 45.77 ± 0.46 (+5.00) & 06.17 ± 0.12 (+1.30) & 03.53 ± 0.15 (+0.56) & 03.60 ± 0.10 (+0.63) \\

        \cmidrule(lr){2-8}
        {} & \textbf{ModPrompt} &  \textbf{93.73 ± 0.25 (+0.60)} & \textbf{68.27 ± 0.40 (+1.10)}  & \textbf{60.87 ± 0.50 (+0.77)} & \textbf{26.63 ± 0.38 (+4.93)} & \textbf{18.53 ± 0.32 (+3.50)} & \textbf{17.27 ± 0.46 (+3.14)} \\

        \bottomrule

    \end{tabular}
    }
    \caption{Detection performance (APs) for YOLO-World and Grounding DINO on LLVIP-IR and NYU$_{v2}$-Depth datasets. Each visual prompt adaptation strategy is compared with the learnable MPDR (results in parenthesis are the gain with the MPDR module), which is responsible for updating the new modality embeddings and not changing the original embedding knowledge.}
    \label{tab:text_residual_study}
\end{table*}

\subsection{Experimental Methodology}

\noindent \textbf{(a) Datasets:} 
\textbf{LLVIP:} LLVIP is a surveillance dataset composed of $12,025$ IR images in the training set and $3,463$ IR images in the test set with only pedestrians annotated. \textbf{FLIR ALIGNED:} We used the sanitized and aligned paired sets provided by Zhang et al.~\cite{zhang2020multispectral}. It has $4,129$ training IR images and $1,013$ IR test images. FLIR images are captured from the perspective of a camera in the front of a car, with a resolution of $640$ by $512$. It contains the bicycles, dogs, cars, and people classes. Following~\citet{cao2023multimodal}, we decided to remove  ``dog" objects because the number of annotations is inadequate for training with the FLIR dataset, results of FLIR can be seen in supp. material due to space constraints. \textbf{NYU$_{v2}$:}. The NYU-Depth V2~\citep{silberman2012indoor} dataset provides a valuable collection of indoor video sequences captured with a Microsoft Kinect camera. In this dataset, we used the depth information, which is composed of $795$ training images and $654$ test images with a resolution of $640$ by $480$. The dataset comprised $19$ different classes, such as bathtubs, beds, bookshelves, boxes, and chairs. In all our experiments, only IR and depth images were used. 

\noindent \textbf{(b) Implementation Details:} 
The YOLO-World models were trained on an A100 NVIDIA GPU and Grounding DINO on V100 and were implemented using PyTorch. For YOLO-World, we used the original work~\citep{cheng2024yolo}, which is based on MMDET~\citep{mmdetection}, and for Grounding DINO, we used the open-source version from the MMDET library. For the YOLO-World, we use both YOLO-World-S-v1 and YOLO-World-S-v2, with the YOLOv8 backbone and CLIP~\citep{radford2021learning} as the language model. For Grounding DINO, we use Swin-Transformer Tiny~\citep{liu2021swin} as the detector backbone and BERT~\citep{devlin2018bert} for the language encoder. Our models are evaluated in terms of APs performance using COCO API~\citep{lin2014microsoft}. The batch size for YOLO-World models was $8$, and the models were trained with a single GPU. For Grounding DINO, the batch size was $16$, and $2$ GPUs were used for training using DDP. Yolo-World and Grounding Dino were trained with the maximum training epoch of $80$ and $60$, respectively. For offline embedding extracting, we used CLIP-ViT-base-patch32 and Bert-base-uncased for YOLO-World and Grounding DINO, respectively. Additional details are provided in suppl. materials.

\noindent \textbf{(c) Baseline Methods:}\\
\noindent  \textbf{- Zero-shot Model.} The zero-shot model refers to performing the inference with the original pre-trained open-vocabulary detector model but on our final detection modality. Normally, these models are trained with large RGB datasets, such as Object365~\citep{O365} or GoldG~\citep{goldg}.\\
\noindent \textbf{- Finetuning.} The finetuning strategy refers to adapting a pre-trained model by changing its parameters directly for the final task, but it destroys the pre-training knowledge of the model. The most common strategies for adapting detectors are the \textit{head finetuning}, which adapts only the head of the detector, and \textit{full finetuning}, which adapts all the weights of the detector.\\
\noindent \textbf{- Visual Prompt.} For this baseline, we define a family of different visual prompt strategies explored in our work. Specifically, we have \textit{fixed}, \textit{random}, \textit{padding}, and \textit{weight map}. In \textit{fixed}, we add a fixed patch in a fixed position for all training. In \textit{random}, we have a random padding that is sampled per image. In \textit{padding}, we add a padding prompt and combine it with the image. In \textit{weight map}, we consider the learnable weights to be a mask with the dimensions of the input image.\\
\noindent \textbf{- Modality Prompt.} Here, we refer to our proposed encoder-decoder visual prompt, in which we learn how to translate the input, guided by the final detection loss, and then we combine its output with the original input.

\begin{table}[!ht]
    \centering
    \resizebox{1.0\columnwidth}{!}{%
    \begin{tabular}{lcqgg}
        \toprule
        \multirow{2}{*}[-0.7em]{\textbf{Method}} & \multirow{2}{*}[-0.7em]{\textbf{Variation}} &   \multicolumn{3}{c}{\textbf{LLVIP - IR}} \\
        \cmidrule(lr){3-5}
        \addlinespace[5pt]        
        {} & {} & \multicolumn{1}{c}{\multirow{2}{*}[1em]{\textbf{AP$_{50}$}}} & 
        \multicolumn{1}{c}{\multirow{2}{*}[1em]{\textbf{AP$_{75}$}}} &
        \multicolumn{1}{c}{\multirow{2}{*}[1em]{\textbf{AP}}}   \\
        \midrule

        \rowcolor{white}
        \multirow{2}{*}[0em]{\textbf{Fixed}}        & 30     & 61.60 ± 0.75 & 39.93 ± 0.52 & 37.97 ± 0.56 \\
        \cmidrule(lr){2-5}
        {}                                 & 300    & 70.30 ± 7.89 & 45.67 ± 6.97 & 43.53 ± 5.79 \\
        \midrule
        \rowcolor{white}
        \multirow{2}{*}[0em]{\textbf{Random}}       & 30     & 60.13 ± 0.29 & 38.73 ± 0.17 & 36.87 ± 0.12 \\
        \cmidrule(lr){2-5}
        {}                                 & 300    & 56.27 ± 0.46 & 33.73 ± 0.62 & 33.13 ± 0.42 \\
        \midrule
        \rowcolor{white}
        \multirow{2}{*}[0em]{\textbf{Padding}}      & 30     & 79.87 ± 1.00 & 51.77 ± 0.90 & 49.30 ± 0.83 \\
        \cmidrule(lr){2-5}
        {}                                 & 300    & 39.53 ± 2.36 & 15.90 ± 1.02 & 19.07 ± 1.18 \\
        
        \midrule
        \rowcolor{white}
        \multirow{2}{*}[0em]{\textbf{ModPrompt}}    & MB     & \textbf{92.80 ± 0.29} & \textbf{70.73 ± 1.02} & \textbf{62.87 ± 0.63} \\
        \cmidrule(lr){2-5}
        {}                                 & RES    & 91.03 ± 0.12 & 68.40 ± 1.10 & 61.43 ± 0.58 \\
                
        \midrule
        \addlinespace[2pt]        
        \midrule

        
        
        
        \multirow{2}{*}[-0.7em]{\textbf{Method}} & \multirow{2}{*}[-0.7em]{\textbf{Variation}} &   \multicolumn{3}{c}{\textbf{NYU$_{v2}$ - {}Depth}} \\
        \cmidrule(lr){3-5}
        \addlinespace[5pt]        
        {} & {} & \multicolumn{1}{c}{\multirow{2}{*}[1em]{\textbf{AP$_{50}$}}} & 
        \multicolumn{1}{c}{\multirow{2}{*}[1em]{\textbf{AP$_{75}$}}} &
        \multicolumn{1}{c}{\multirow{2}{*}[1em]{\textbf{AP}}}   \\
        \midrule
        \rowcolor{white}
        \multirow{2}{*}[0em]{\textbf{Fixed}}        & 30     & 04.67 ± 0.05 &  03.07 ± 0.05 &  02.90 ± 0.00 \\
        \cmidrule(lr){2-5}
        {}                                 & 300    & 03.43 ± 0.05 &  02.00 ± 0.08 &  02.10 ± 0.00 \\
        \midrule
        \rowcolor{white}
        \multirow{2}{*}[0em]{\textbf{Random}}       & 30     & 04.23 ± 0.12 &  02.63 ± 0.05 &  02.53 ± 0.05 \\
        \cmidrule(lr){2-5}
        {}                                 & 300    & 01.53 ± 0.17 &  00.77 ± 0.12 &  00.87 ± 0.12 \\
        \midrule
        \rowcolor{white}
        \multirow{2}{*}[0em]{\textbf{Padding}}      & 30     & 03.97 ± 0.05 &  02.50 ± 0.00 &  02.43 ± 0.05 \\
        \cmidrule(lr){2-5}
        {}                                 & 200    & 00.37 ± 0.12 &  00.10 ± 0.08 &  00.17 ± 0.05 \\
        
        \midrule
        \rowcolor{white}
        \multirow{2}{*}[0em]{\textbf{ModPrompt}}    & MB     & 35.37 ± 0.12 &  25.20 ± 0.24 &  23.27 ± 0.17 \\
        \cmidrule(lr){2-5}
        {}                                 & RES    & \textbf{37.17 ± 0.57} &  \textbf{27.50 ± 0.64} &  \textbf{24.93 ± 0.50} \\

        \bottomrule
    \end{tabular}
    }
    \caption{Detection performance (APs) for YOLO-World under the two main datasets evaluated: LLVIP-IR and NYU$_{v2}$-Depth. We compared the main visual prompt strategies \textit{fixed}, \textit{random}, \textit{padding}, and ModPrompt. The variations consist of the number of prompt pixels (p$_{s}=30$, $200$ or $300$) and for ModPrompt, the MobileNet (MB) or ResNet (RES).}
    \label{tab:ablation_tab}
    \vspace{-.6cm}
\end{table}

\subsection{Visual Modality Adaptation}

This section explores two main open-vocabulary detectors, YOLO-World and Grounding DINO, for the input space visual prompt adaptation on IR and depth modalities. In the~\tref{tab:main_yolo_world}, we tabulate the results using YOLO-World and Grounding DINO. We compare zero-shot (ZS), full-finetuning (FT) of the visual backbone, visual prompt with fixed patch, random patch, padding patch, weight map (WM) on all pixels and weight map with scale shift (WM$_{v2}$), and our ModPrompt. Every visual prompt strategy adapts at the input level, so we keep all other parameters frozen to avoid catastrophic forgetting. In this setup, the YOLO-World with LLVIP-IR dataset, we observe that ModPrompt outperforms the other prompt strategies with AP$_{50}$ $92.80$ and the second best was WM with $82.00$, similar trends for the AP$_{75}$ with ModPrompt reaching $70.73$ and WM with $53.90$. Additionally, in terms of AP, our method has $62.87$ AP, while the second best approach has  $50.90$ AP. Here, it is important to mention that when designing visual prompt strategies that work at the input level, the prompt can even be worse for adapting to a new modality when compared with zero-shot for OD performance. For Grounding DINO, in LLVIP-IR, our ModPrompt had a $93.13$ AP$_{50}$, $67.17$ AP$_{75}$ and $60.10$ AP and the second best (fixed patch) had an AP$_{50}$ of $83.87$ but close to random and padding prompts. We observe that ModPrompt performs better when objects are well-defined in the image and when objects are not too small, otherwise, like all other input-level pixel strategies it struggles, especially on refined bounding-box localization, which can be seen with AP$_{75}$ and AP, whereas in AP$_{50}$ it shows good results. Performance on the FLIR dataset is reported in the supp. material. \newline

\begin{figure}
\centering
\begin{subfigure}[t]{1.0\columnwidth}

    \includegraphics[width=\textwidth]{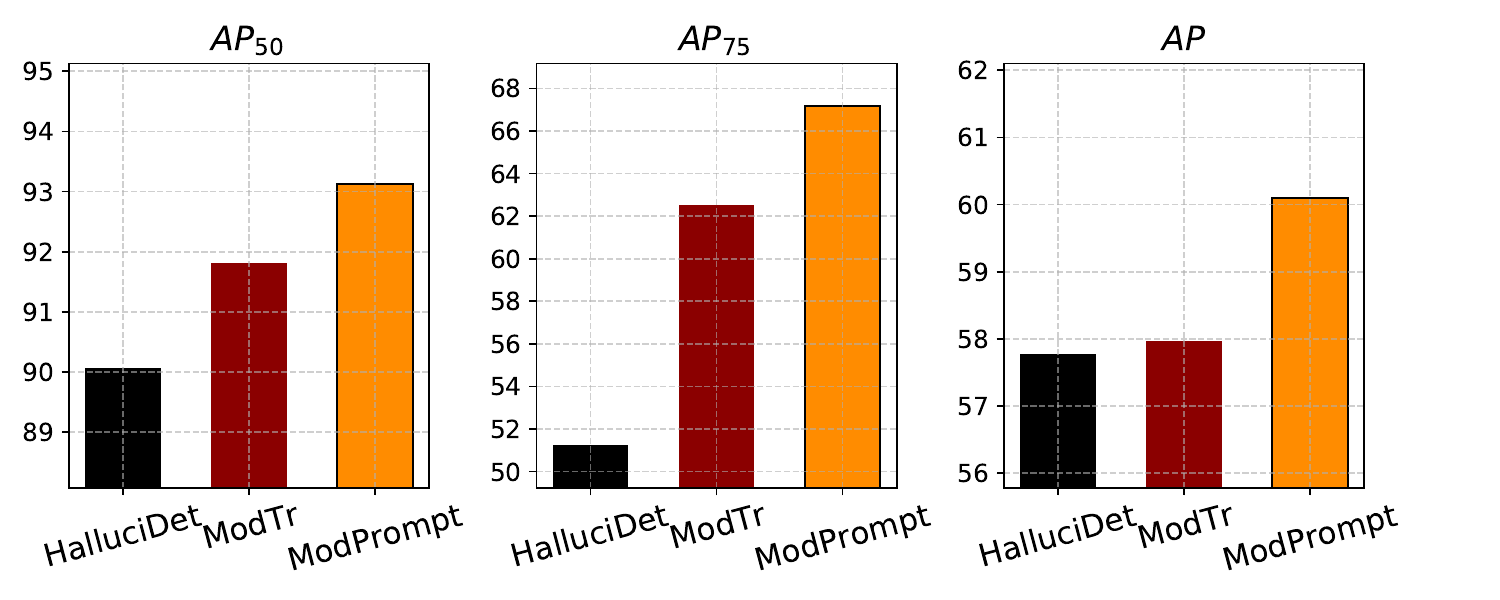}
\end{subfigure}

\caption{Detection performance on LLVIP dataset of different SOTA Modality Translation OD methods in terms of APs.}
\label{fig:comparative}
\vspace{-.4cm}
\end{figure}

\noindent \textbf{Comparison with SOTA Modality Translation OD methods.} Our ModPrompt technique is compared with recent works on modality translators for ODs: HalluciDet~\citep{medeiros2024hallucidet} and ModTr~\citep{medeiros2024modality}. In~\cref{fig:comparative}, we observe that our results are better in all APs for the LLVIP dataset. Additional results for the FLIR-IR dataset are presented in the supplementary material. \newline

\noindent \textbf{Comparison of the number of Train Parameters and Knowledge Preservation.} In~\tref{tab:comparison_knowledgepres} we compare the number of training parameters and show the catastrophic forgetting in the compared baseline methods. As we can see, ModPrompt preserves the zero-shot performance on the COCO dataset while the performance drops significantly in HFT and FT. Other visual prompt strategies such as WM and WM$_{2}$ can preserve the zero-shot performance, but our average performance is much better than theirs while requiring less trainable parameters.


\begin{table}[!ht]
    \centering
    \resizebox{1.0\columnwidth}{!}{%
    \begin{tabular}{qggggg} 
        \toprule
        \rowcolor{white}
        
        \multirow{2}{*}[+0.5em]{\textbf{Method}} & \multicolumn{1}{c}{\textbf{Params (M)}} & \multicolumn{1}{c}{\textbf{LLVIP}} &  \multicolumn{1}{c}{\textbf{COCO}} & \multicolumn{1}{c}{\textbf{Avg.}} \\
 
        \rowcolor{white}
        \midrule
        
        ZS  & 0.00 & 81.00 ± 0.00  & 51.90 ± 0.00  & 66.45 (00.00) \\
        HFT & 2.31 & 93.57 ± 0.05  & 00.66 ± 0.04  & 47.12 (-19.33) \\
                          
        \rowcolor{white}
        FT  & 76.81 & 97.43 ± 0.05  & 00.10 ± 0.00  & 48.77 (-17.68) \\
                           
        \midrule
        WM & 3.93 & 87.47 ± 0.17  & 51.90 ± 0.00  & 69.69 (+3.24) \\
        
        \rowcolor{white}
        WM$_{v2}$ & 7.86 & 85.00 ± 0.08  & 51.90 ± 0.00  & 68.45 (+2.00) \\

        \midrule
        \textbf{ModPrompt} & 3.08 & 95.63 ± 0.04  & 51.90 ± 0.00  & \textbf{73.77 (+7.32)} \\
 
        \bottomrule
    \end{tabular}
    }
    \caption{AP$_{50}$ of YOLO-World on LLVIP-IR and COCO data. We compare the number of trainable parameters and show the catastrophic forgetting in HFT and FT baselines.}
    \label{tab:comparison_knowledgepres}
    \vspace{-.4cm}
\end{table}

\begin{figure*}
\centering
\begin{subfigure}[t]{0.15\textwidth}
    \caption{GT}

    \makebox[0pt][r]{\makebox[15pt]{\raisebox{40pt}{\rotatebox[origin=c]{90}{}}}}%
    \includegraphics[width=\textwidth]{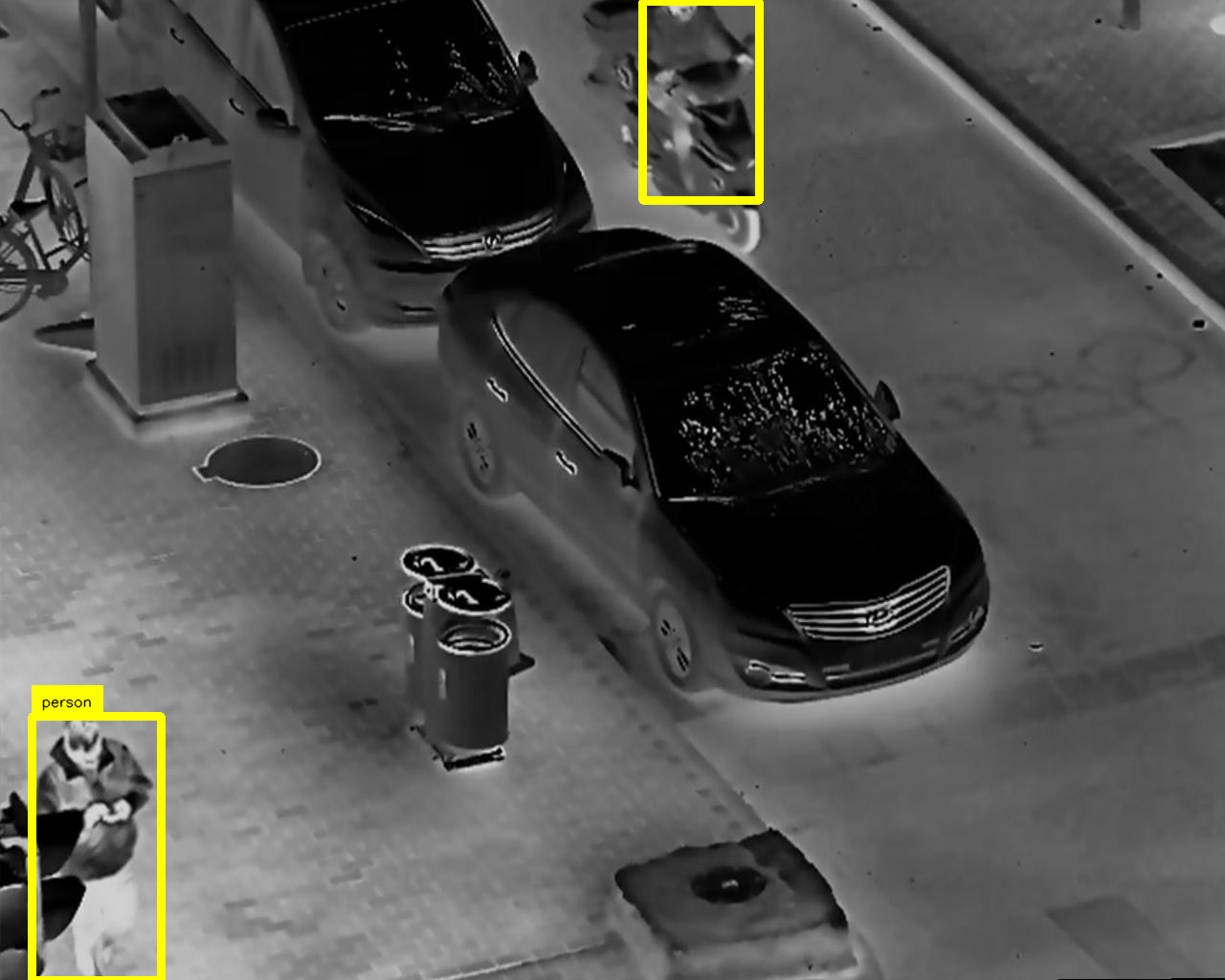}

    \makebox[0pt][r]{\makebox[15pt]{\raisebox{40pt}{\rotatebox[origin=c]{90}{}}}}%
    \includegraphics[width=\textwidth]{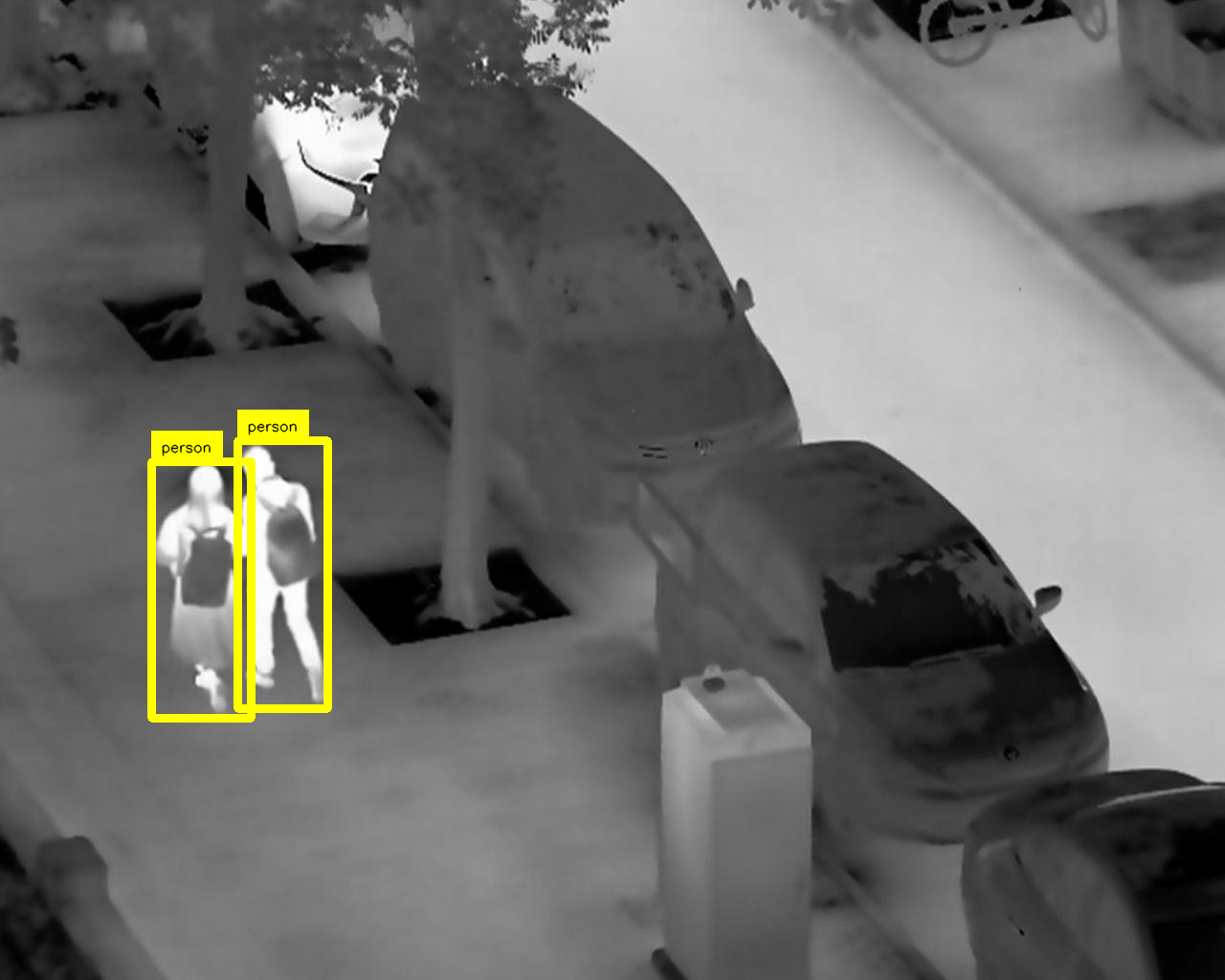}

    \makebox[0pt][r]{\makebox[15pt]{\raisebox{40pt}{\rotatebox[origin=c]{90}{}}}}%
    \includegraphics[width=\textwidth]{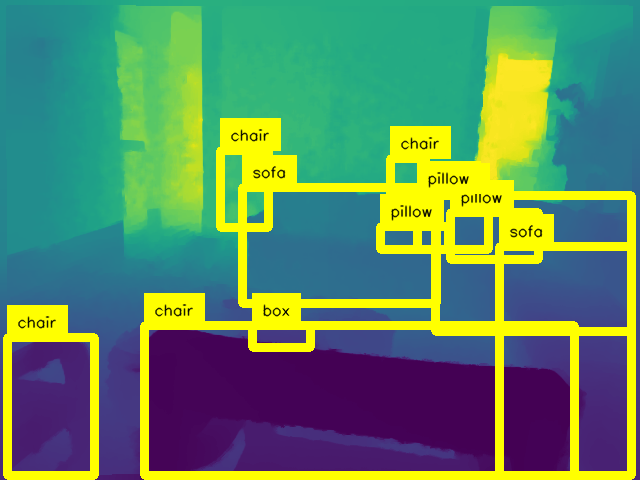}

    \makebox[0pt][r]{\makebox[15pt]{\raisebox{40pt}{\rotatebox[origin=c]{90}{}}}}%
    \includegraphics[width=\textwidth]{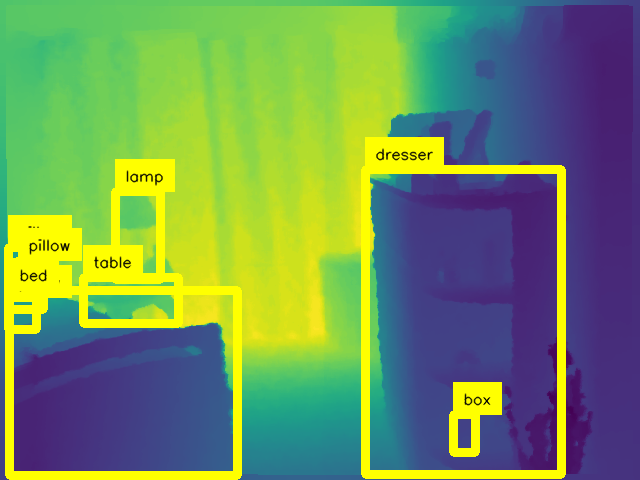}

\end{subfigure}
\begin{subfigure}[t]{0.15\textwidth}
    \caption{Zero-Shot}

    \includegraphics[width=\textwidth]{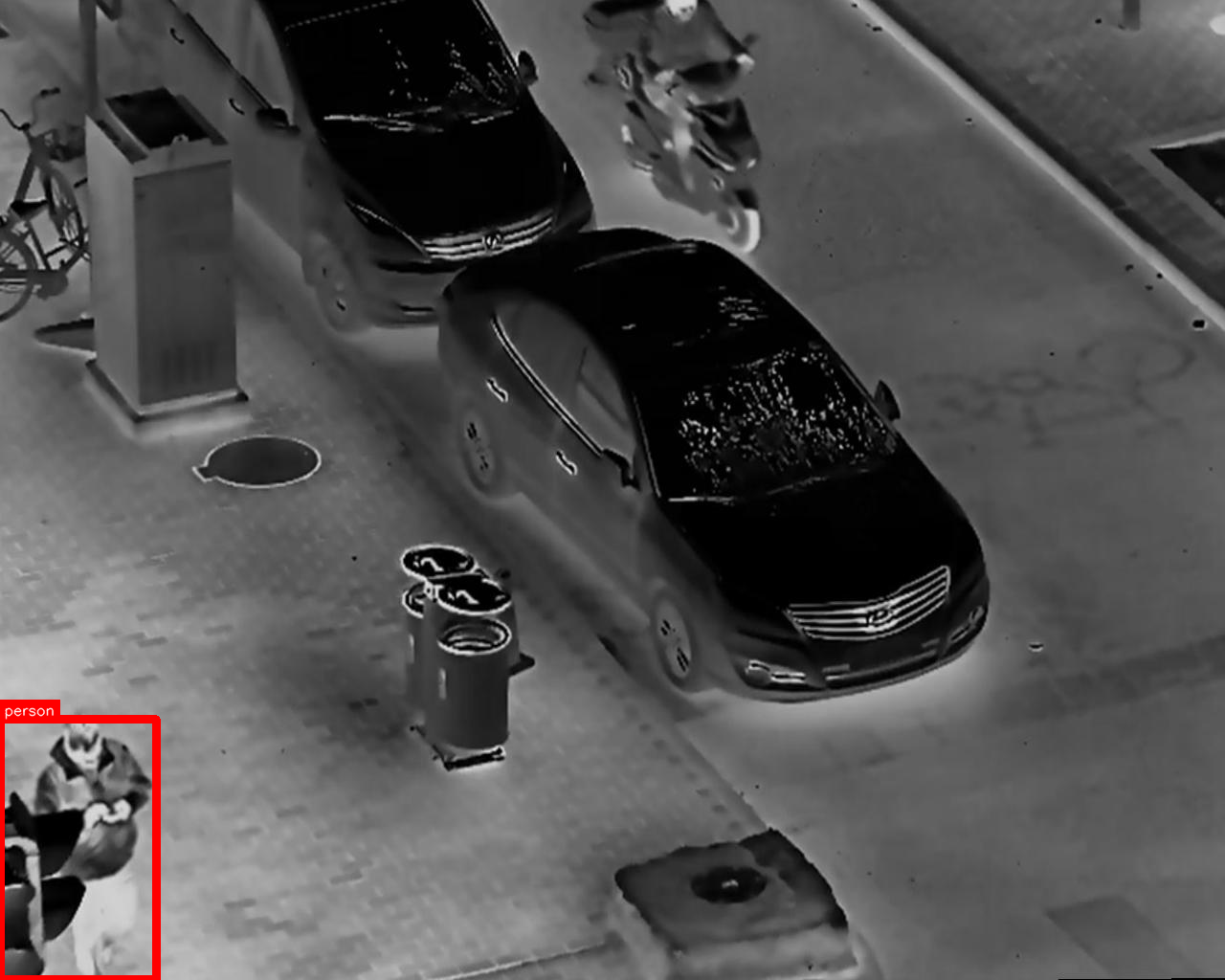}

    \includegraphics[width=\textwidth]{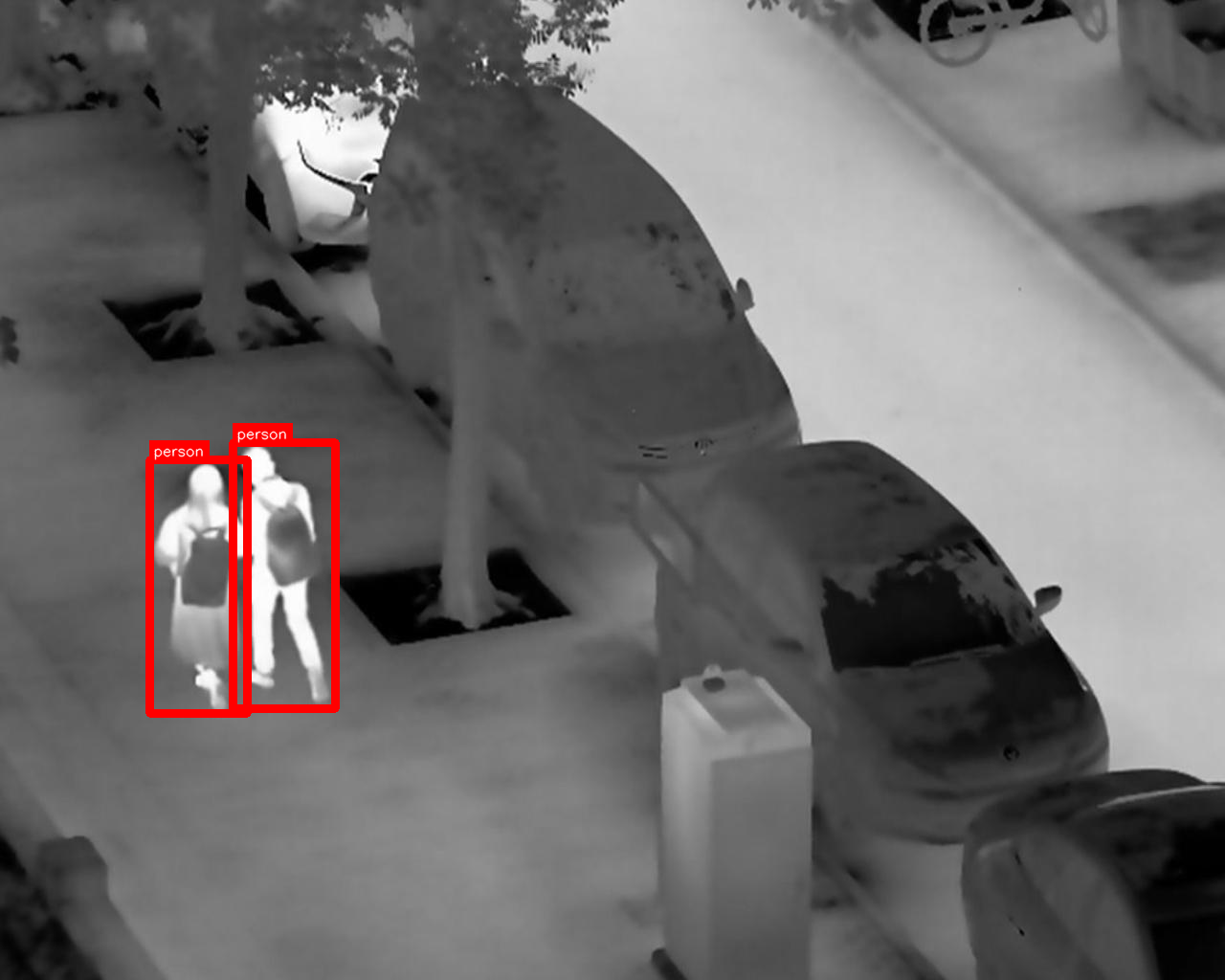}
    
    \includegraphics[width=\textwidth]{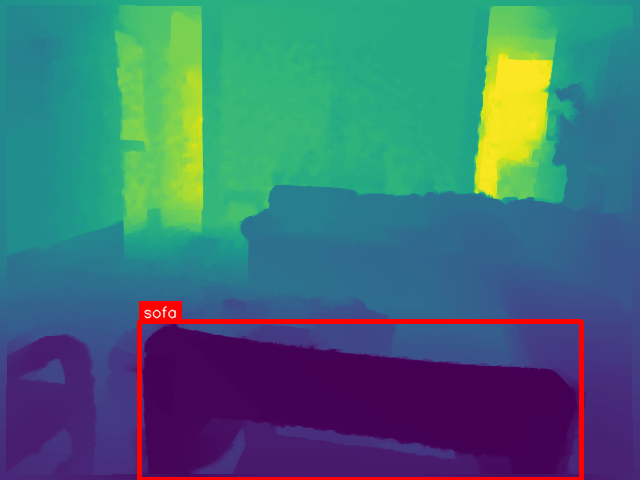}

    \includegraphics[width=\textwidth]{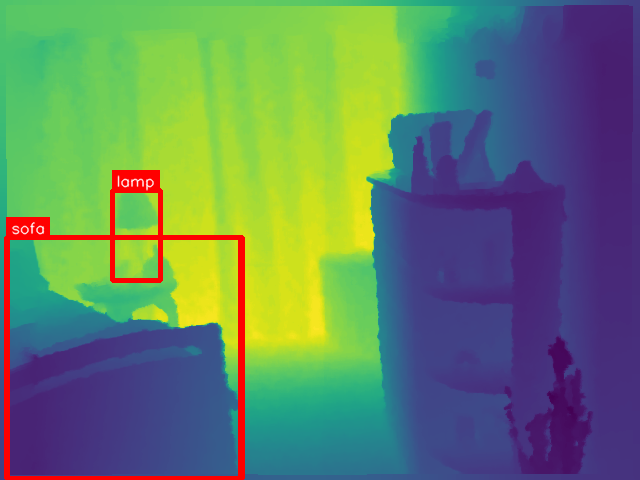}

\end{subfigure}
\begin{subfigure}[t]{0.15\textwidth}
    \caption{Visual Prompt}
    
    \includegraphics[width=\textwidth]{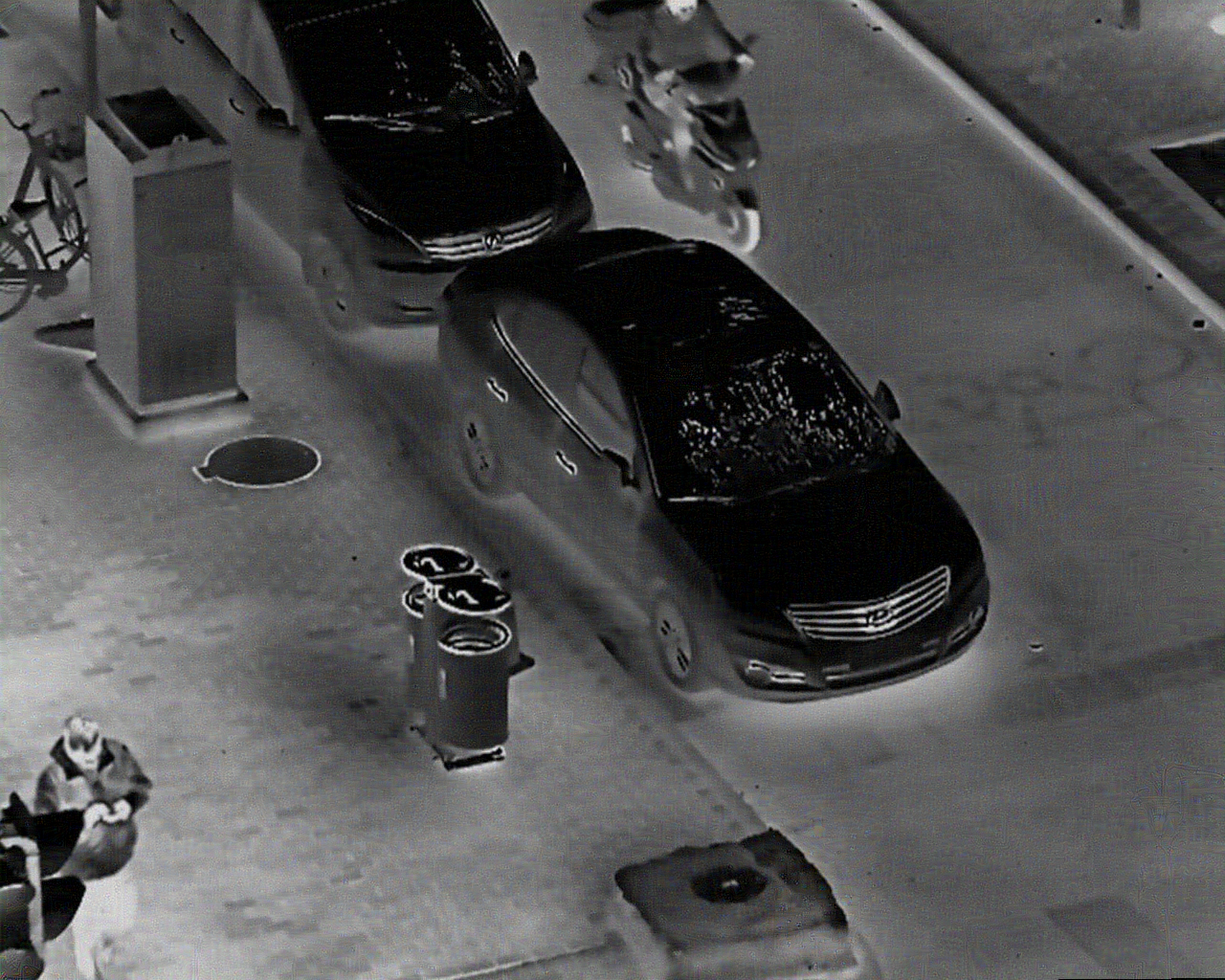}
    
    \includegraphics[width=\textwidth]{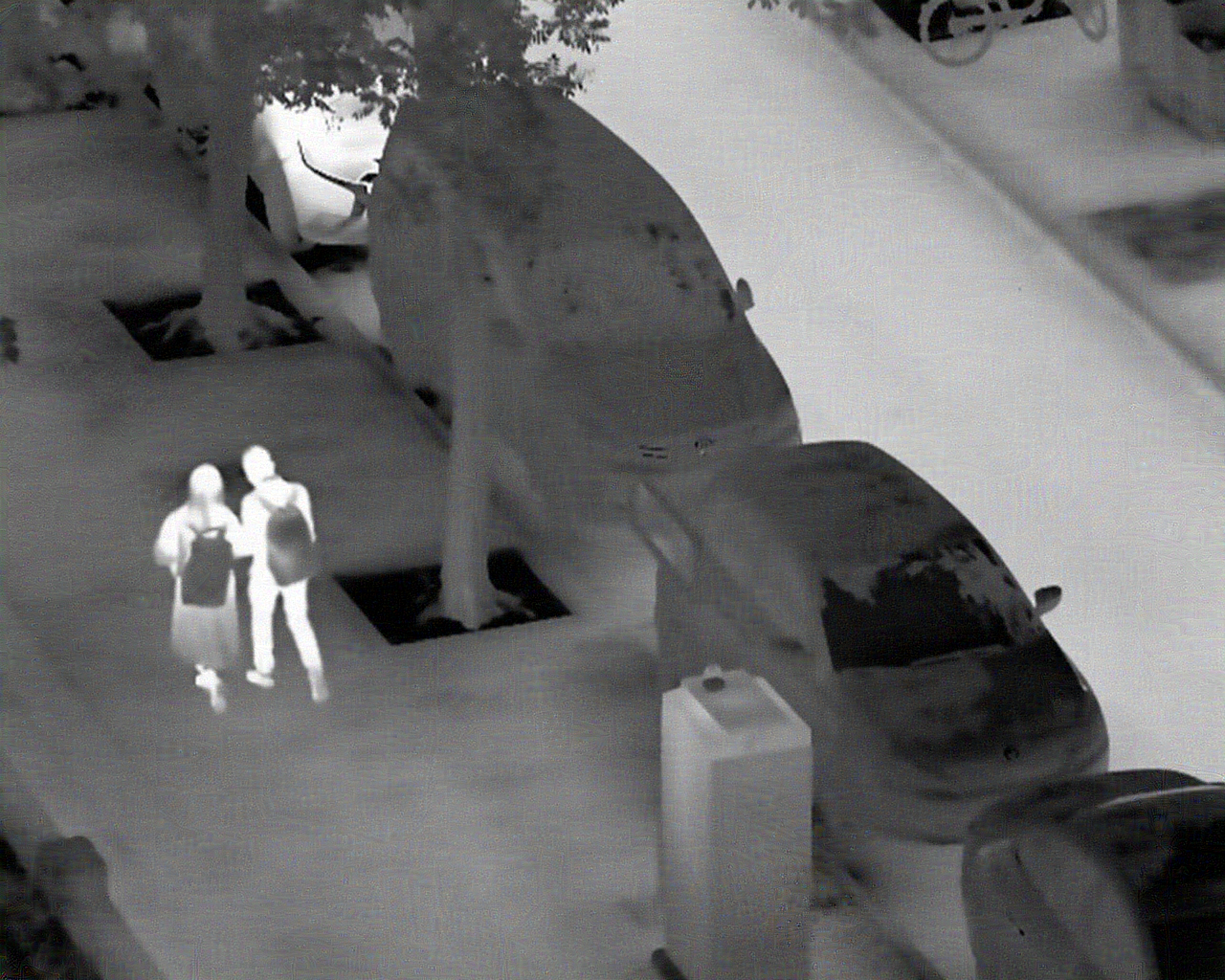}
    
    \includegraphics[width=\textwidth]{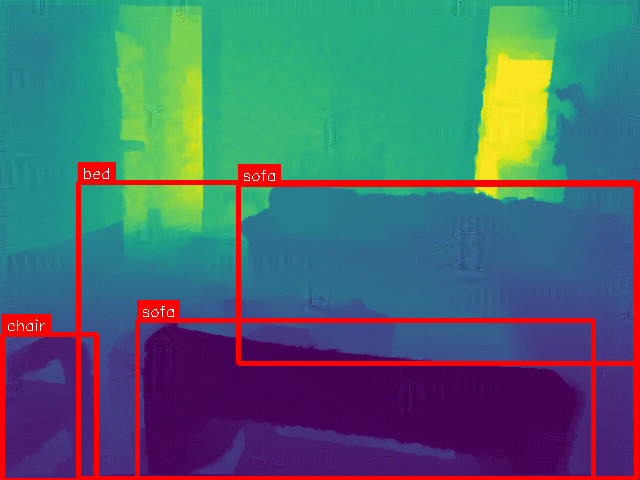}
    
    \includegraphics[width=\textwidth]{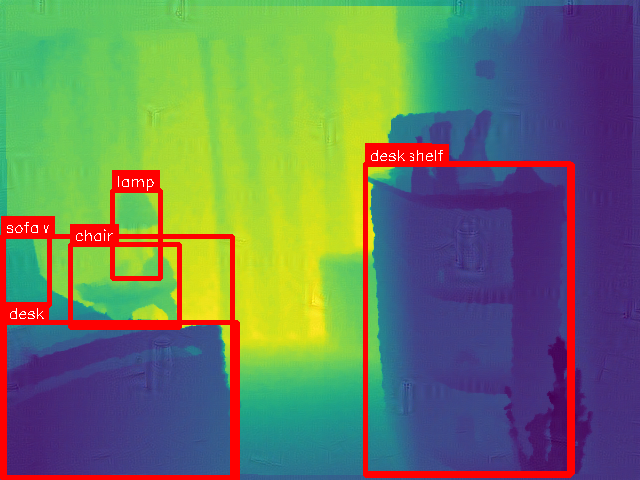}
    
\end{subfigure}
\begin{subfigure}[t]{0.15\textwidth}
    \caption{ModPrompt}

    \includegraphics[width=\textwidth]{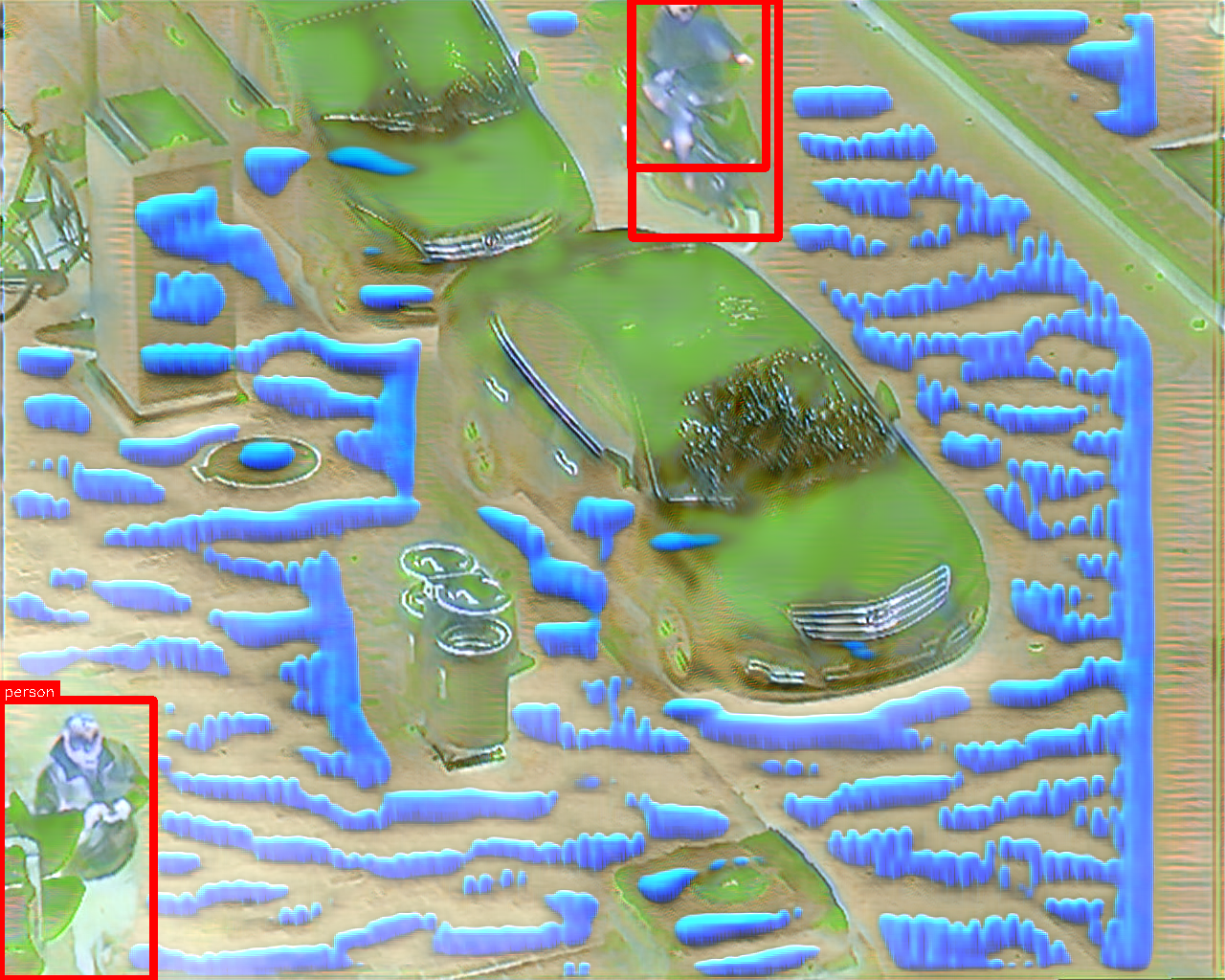}
    
    \includegraphics[width=\textwidth]{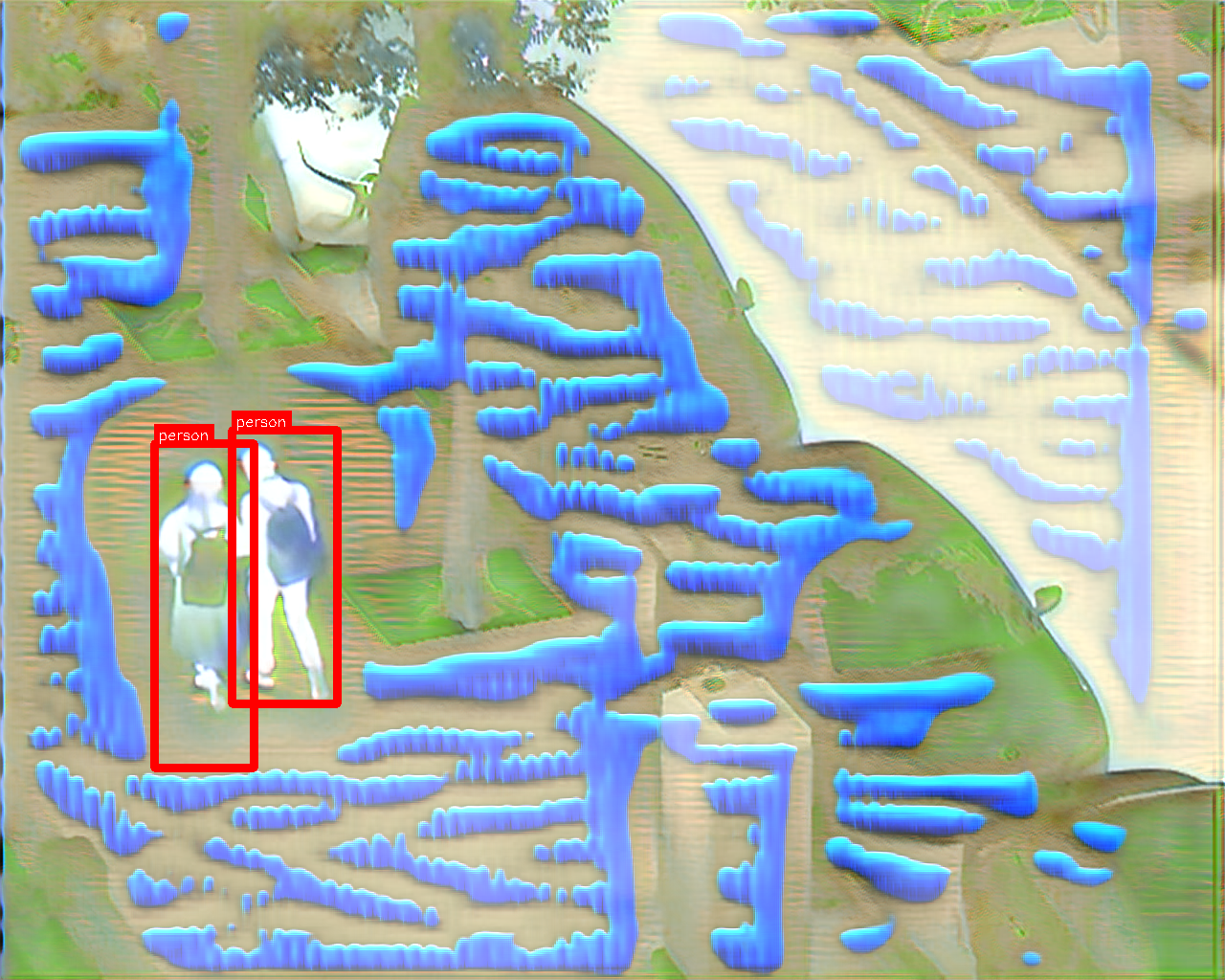}
    
    \includegraphics[width=\textwidth]{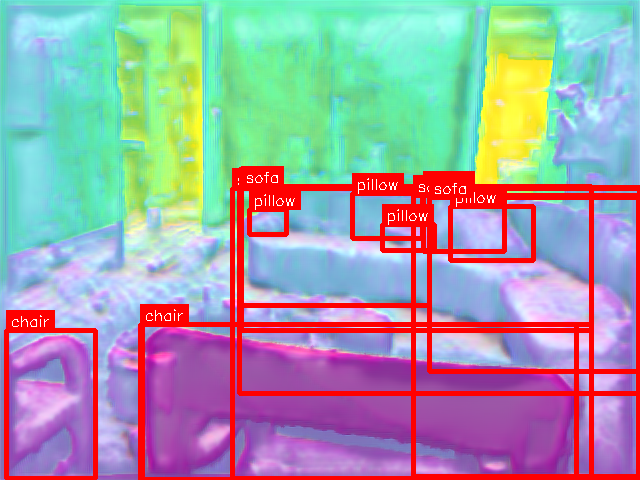}
    
    \includegraphics[width=\textwidth]{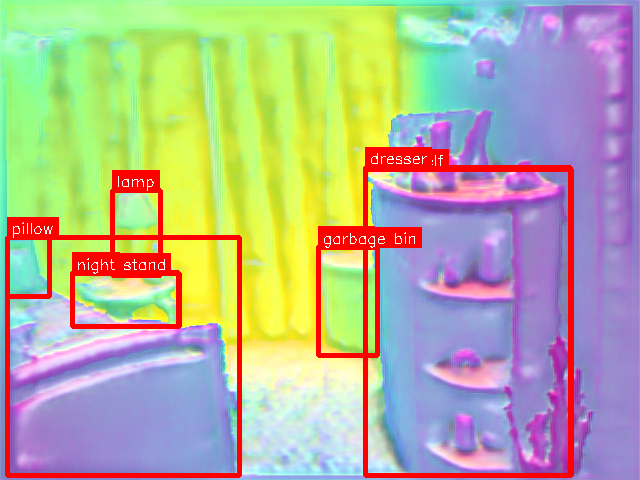}
    
\end{subfigure}

\caption{\textbf{Detections for YOLO-World for the different approaches}: First two rows for LLVIP (infrared), and last two rows for NYU$_{v2}$ (depth). Each column corresponds to a different approach: \textbf{(a) GT (Ground Truth):} Shows in yellow the ground-truth bounding boxes for objects. \textbf{(b) Zero-Shot:} Displays detections (in red) from a zero-shot model. \textbf{(c) Visual Prompt:} Illustrates detections with a visual prompt added to the image. \textbf{(d) ModPrompt (Ours):} Detections from our proposed model.}
\label{fig:qualitative_results}
\vspace{-.3cm}
\end{figure*}

\subsection{Ablation Studies}

\noindent \textbf{Visual Prompts.} 
We evaluate different variations of the visual prompt adaptation methods introduced in our paper. Specifically, we compare the performance when different input patch sizes are used; for instance, p$_{s} = 30$ refers to a patch size of $30$ pixels. In this study, we tested various patch sizes for each of the visual prompt methods and reported the performance in~\tref{tab:ablation_tab}. We evaluate modprompt using two different translators with U-Net based backbones- MobileNet (MB)~\citep{howard2017mobilenets} and ResNet (RES)~\citep{he2016deep}. Empirically, we observe minimal variation in results when increasing the prompt patch size. In some cases, a larger prompt could even degrade performance of the detector on the new modality, as seen with p$_{s} = 200$ on NYU$_{v2}$ datasets. A similar trend can be observed with ModPrompt, where sometimes the MB version, with far fewer parameters, can outperform the RES version. This finding suggests that a lighter MB translator can be a viable choice over a heavier RES translator, making it better suited for real-time performance, with minimal degradation in detection metrics. \newline

\noindent \textbf{MPDR Knowledge Preservation mechanism.} 
\tref{tab:text_residual_study} shows that MDPR can be beneficial for improving the majority of the visual prompt strategies while preserving the original knowledge of the detector. For the text adaptation setting, we pre-compute the embeddings for the target classes using our language model, avoiding forward/backpropagation on it. In this setting, we also tried performing adaptations directly on the embeddings without using MPDR parameters, similar to text/embedding prompt tuning; in this case, the performance on the final target modality was close to MDPR. However, the zero-shot knowledge of the model was lost, thus we opted to report only the MDPR in the main manuscript.

\subsection{Qualitative Results}

\fref{fig:qualitative_results} shows the visualization of ModPrompt on YOLO-World compared with zero-shot and the best visual prompt baseline. Thus, we can see that visual prompt-based methods, in general, struggle a bit to provide attached bounding boxes, but they do a good job of adapting. The VP method could not detect people in the LLVIP image well, while ModPrompt did a good job. Additional visualizations are provided in the supplementary material.
\section{Conclusion} 
\label{sec:conclusion}

In this work, we presented ModPrompt, a novel visual prompt strategy that effectively adapts open-vocabulary ODs to new visual modalities, such as infrared and depth, while preserving the detector’s zero-shot capabilities. Through our encoder-decoder prompt approach, we achieve significant performance improvements across multiple challenging benchmarks, showcasing the robustness of our method in scenarios with different modality adaptations. Moreover, we introduced MPDR, an efficient text-embedded prompt tuning decoupled from the original embedded knowledge that further enhances detection performance by preserving language knowledge during adaptation, thereby, in some cases, achieving results comparable to full fine-tuning. Our evaluations demonstrate that ModPrompt consistently outperforms existing pixel-level prompt strategies and adapts seamlessly to varied backbone architectures, such as CNN-based or transformer-based. By expanding the potential of vision-language models in different applications, ModPrompt offers a compelling solution for future advancements in adaptable object detection frameworks, especially in settings that require full knowledge preservation.

\section*{Acknowledgments} This work was supported in part by Distech Controls Inc., the Natural Sciences and Engineering Research Council of Canada, the Digital Research Alliance of Canada, and MITACS. \\

\twocolumn[{%
\renewcommand\twocolumn[1][]{#1}%
\maketitlesupplementary
\begin{center}
    \centering
    \captionsetup{type=figure}
    \begin{subfigure}[!htp]{0.24\textwidth}
        \caption{GT}
    
        \makebox[0pt][r]{\makebox[18pt]{\raisebox{40pt}{\rotatebox[origin=c]{90}{LLVIP}}}}%
        \includegraphics[width=\textwidth]{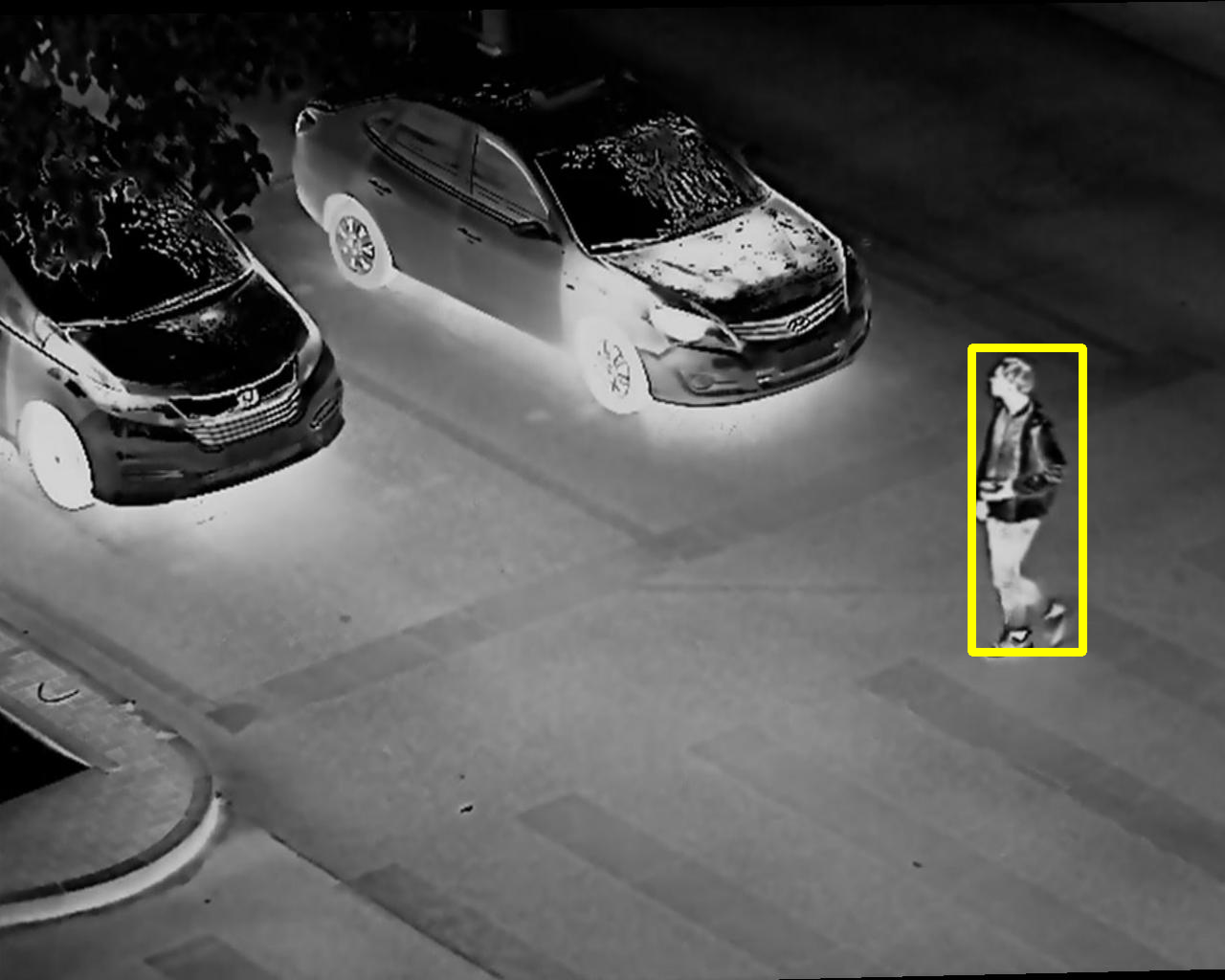}
        
        \makebox[0pt][r]{\makebox[18pt]{\raisebox{40pt}{\rotatebox[origin=c]{90}{FLIR}}}}%
        \includegraphics[width=\textwidth]{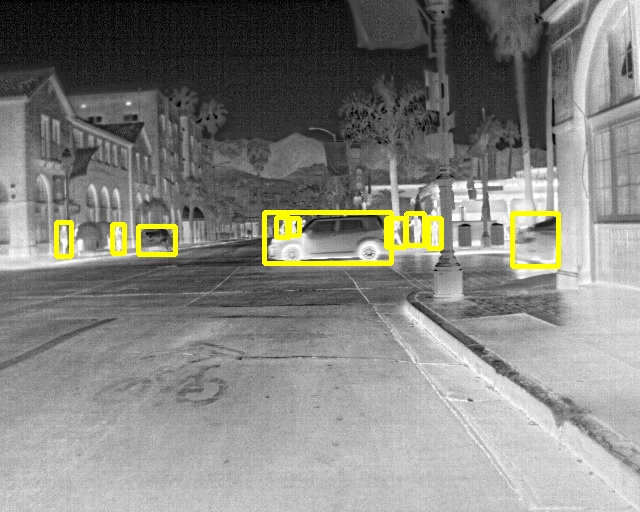}
        
        \makebox[0pt][r]{\makebox[18pt]{\raisebox{40pt}{\rotatebox[origin=c]{90}{NYU$_{v2}$}}}}%
        \includegraphics[width=\textwidth]{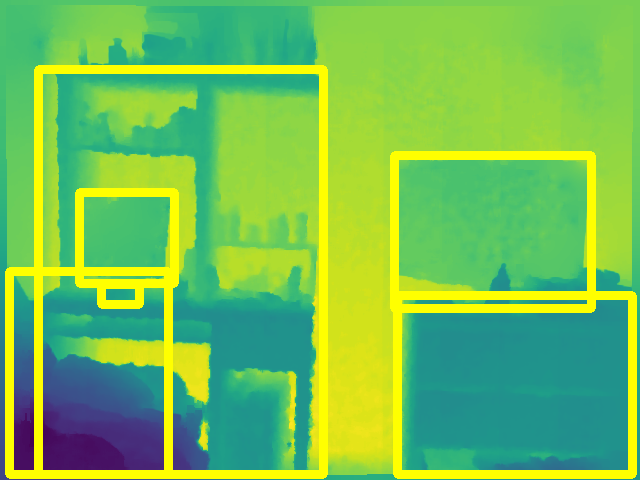}

    \end{subfigure}
    \begin{subfigure}[!htp]{0.24\textwidth}
        \caption{Zero-Shot}
    
        \includegraphics[width=\textwidth]{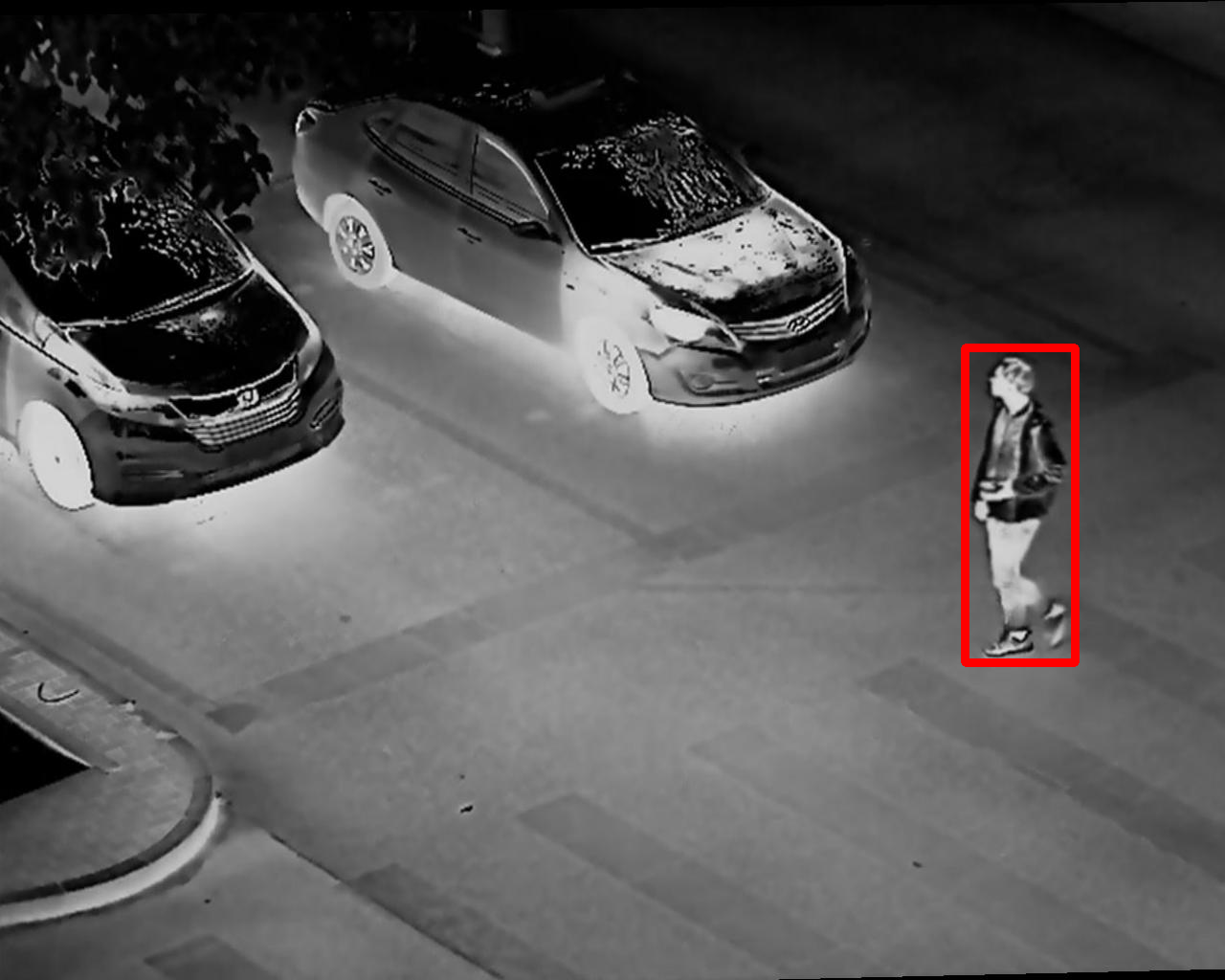}

        \includegraphics[width=\textwidth]{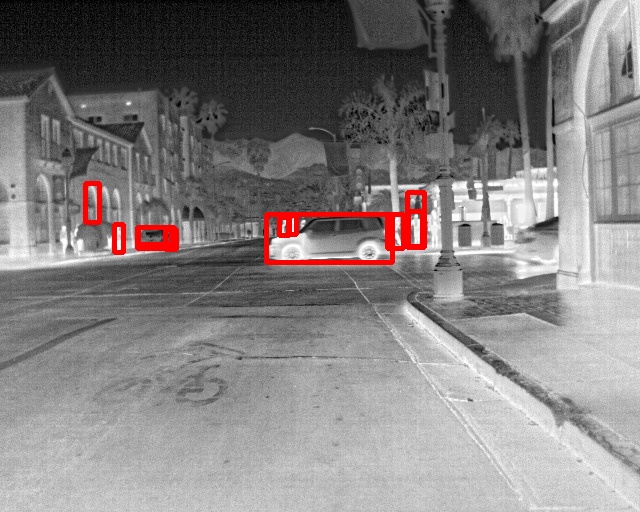}
        
        \includegraphics[width=\textwidth]{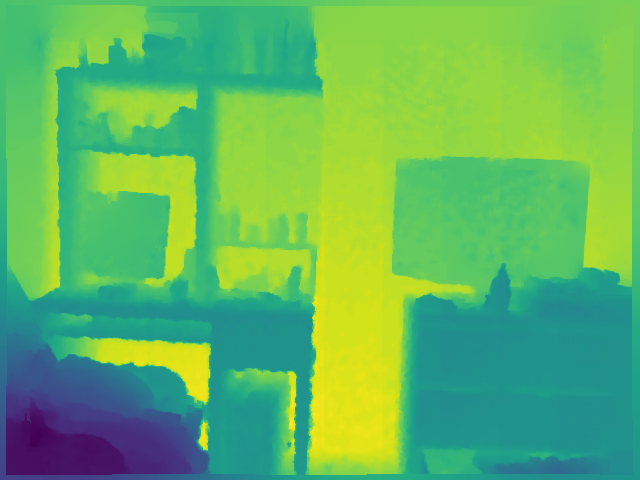}
        
    \end{subfigure}
    \begin{subfigure}[!htp]{0.24\textwidth}
        \caption{Visual Prompt}
        
        \includegraphics[width=\textwidth]{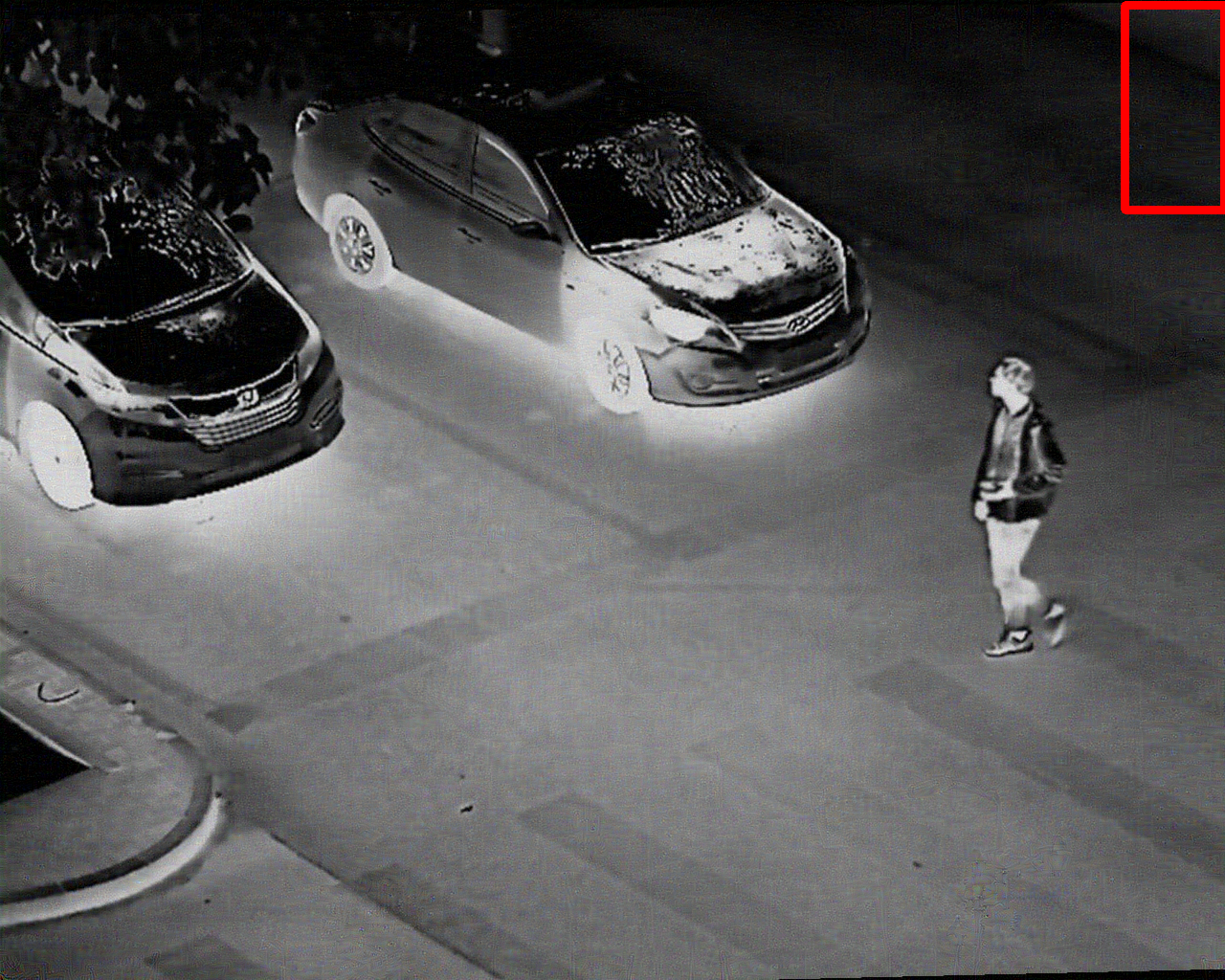}
        
        \includegraphics[width=\textwidth]{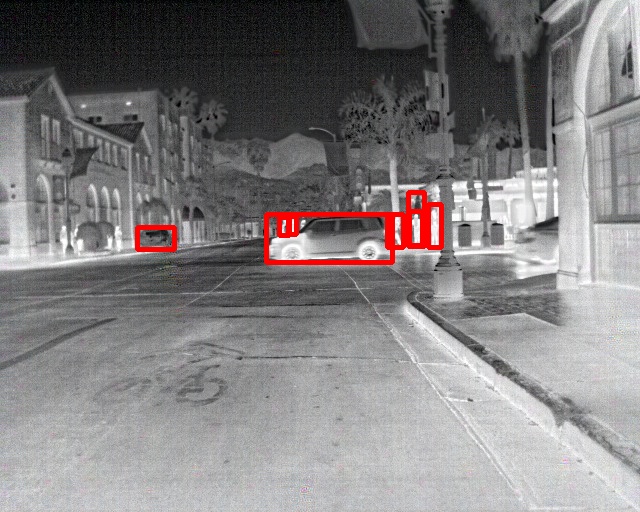}
        
        \includegraphics[width=\textwidth]{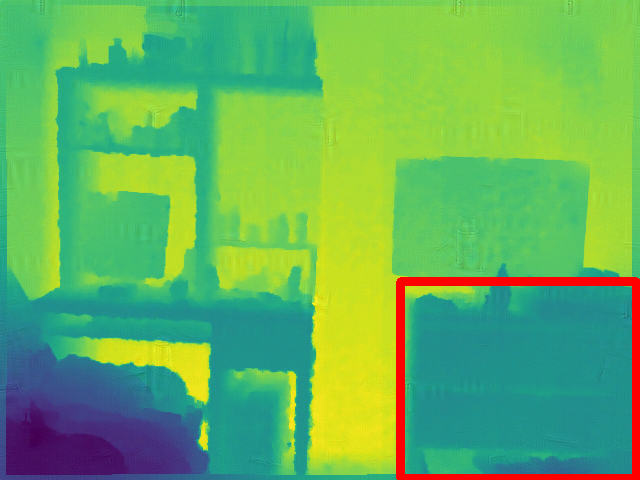}
        
    \end{subfigure}
    \begin{subfigure}[!htp]{0.24\textwidth}
        \caption{ModPrompt (Ours)}
    
        \includegraphics[width=\textwidth]{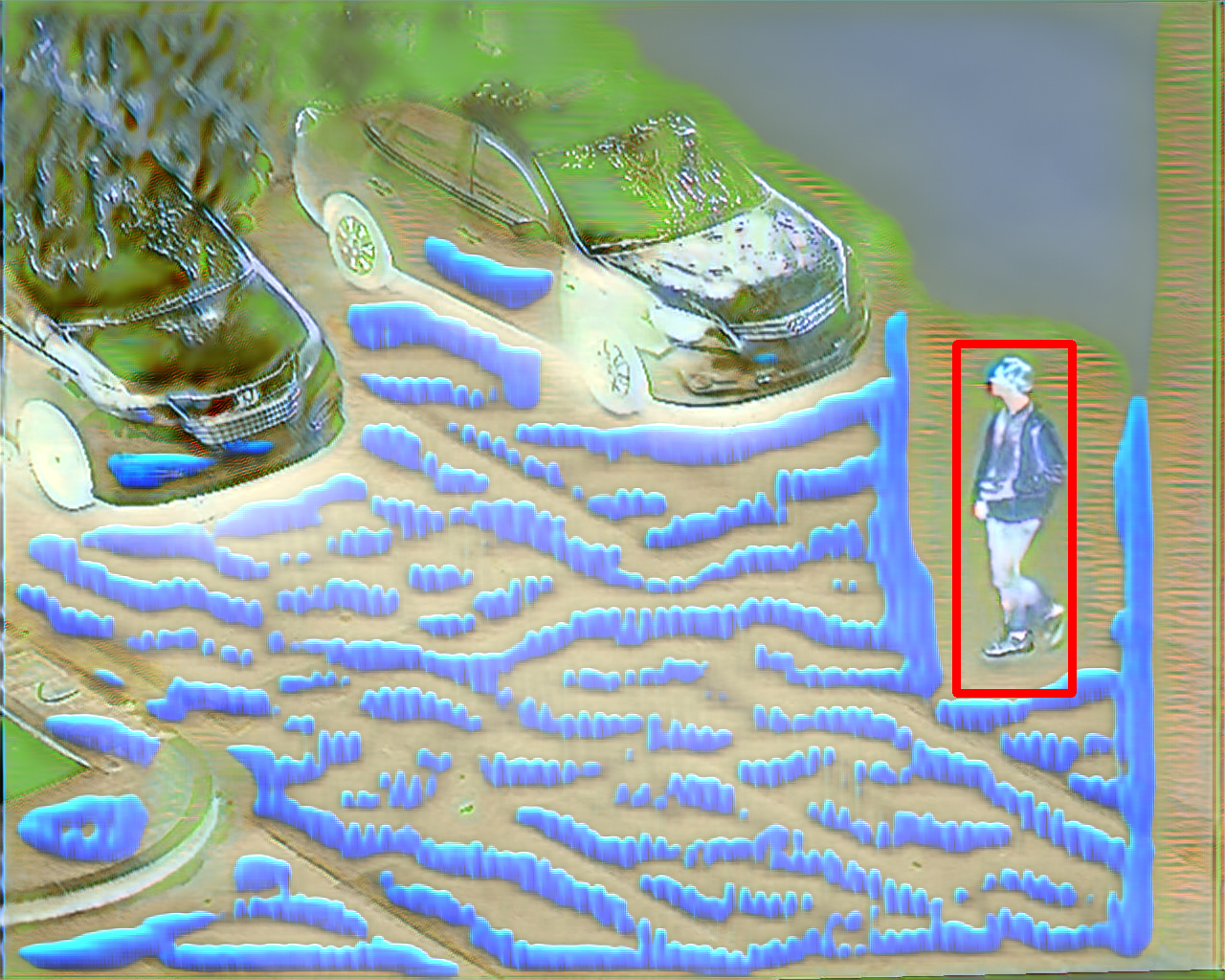}
        
        \includegraphics[width=\textwidth]{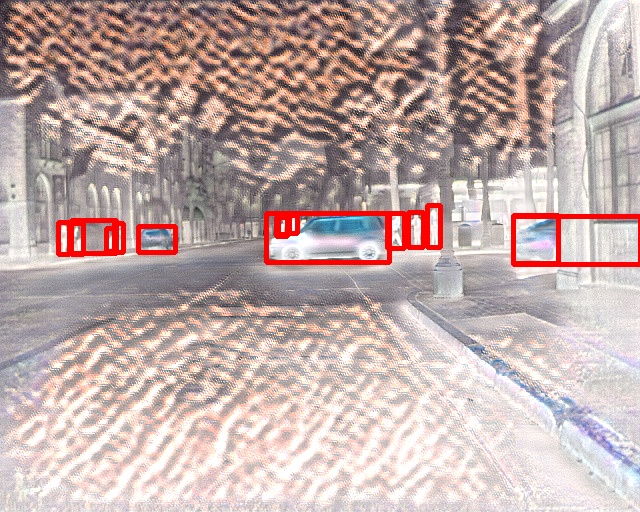}
        
        \includegraphics[width=\textwidth]{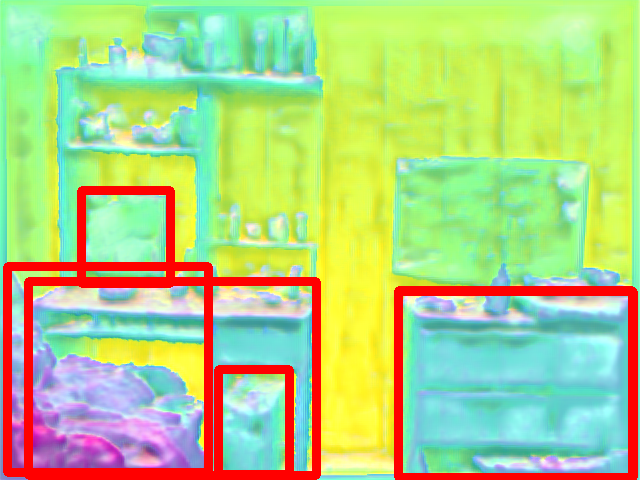}
        
    \end{subfigure}

    \setcounter{figure}{0}
    \captionof{figure}{\textbf{Detections of different approaches across modalities}: LLVIP and FLIR datasets (infrared) and NYU$_{v2}$ (depth). Each column corresponds to a different approach: \textbf{(a) GT (Ground Truth):} Shows in yellow the ground-truth bounding boxes for objects. \textbf{(b) Zero-Shot:} Displays detections (in red) from a zero-shot model. This model misses several detections and predicts inaccurate boxes without specific tuning. \textbf{(c) Visual Prompt:} Illustrates detections with weight map visual prompt added to the image. It shows improvements over zero-shot, with more accurate detection, but still misses some objects. \textbf{(d) ModPrompt (Ours):} Detections from our proposed model. ModPrompt generates artifacts on the image to enhance objects and suppress background, facilitating the detector.
    \label{figure:initial_fig}
}
\end{center}%
}]

In the supplementary material, we provide additional information to reproduce our work. The source code is provided alongside the supplementary material. The supplementary material is divided into the following sections: \sref{sec:additional_details} with additional details regarding the implementation and architecture of the vision-language object detectors used in our work. Then in \sref{sec:formal_definition_tpt} we formally define text-prompt tuning in more detail. And in \sref{sec:results} we provide some additional results. Specifically, in~\sref{sec:results_flir} we provide additional main results on the FLIR-IR dataset. Then, we provide results with different backbones, YOLO-World-Large and Grounding DINO-B in \sref{sec:different_detectors} on FLIR-IR and NYU$_{v2}$-DEPTH. Further, in~\sref{sec:inference_friendly} we provide the results for the FLIR-IR dataset with the learnable MPDR. In~\sref{sec:ablations_vp} we provide results of ablations using different visual prompt strategies. Then, in~\sref{sec:sota} we compare our method with state-of-the-art modality translation OD methods, and in~\sref{sec:additional_qualitative_results} we show additional visualizations. Finally, in~\sref{sec:limitations_futureworks} we provide some limitations of our work and possible future directions.

\begin{table*}[!ht]
    \centering
    \resizebox{1.0\textwidth}{!}{%
    \begin{tabular}{lqgggggg} 
        \toprule
        \rowcolor{white}
        \multirow{2}{*}[-0.7em]{\textbf{Dataset}} & \multirow{2}{*}[-0.7em]{\textbf{Method}} &  \multicolumn{3}{c}{\textbf{YOLO-World}} &  \multicolumn{3}{c}{\textbf{Grounding DINO}} \\
        \cmidrule(lr){3-5}
        \cmidrule(lr){6-8}
        \addlinespace[5pt]   

        \rowcolor{white}
        {} & {} & \multicolumn{1}{c}{\multirow{2}{*}[1em]{\textbf{AP$_{50}$}}} & 
        \multicolumn{1}{c}{\multirow{2}{*}[1em]{\textbf{AP$_{75}$}}} &
        \multicolumn{1}{c}{\multirow{2}{*}[1em]{\textbf{AP}}} & \multicolumn{1}{c}{\multirow{2}{*}[1em]{\textbf{AP$_{50}$}}} & 
        \multicolumn{1}{c}{\multirow{2}{*}[1em]{\textbf{AP$_{75}$}}} &
        \multicolumn{1}{c}{\multirow{2}{*}[1em]{\textbf{AP}}}   \\
        \midrule

        {} & Zero-Shot (ZS)  & 64.70 ± 0.00 & 32.10 ± 0.00 & 34.90 ± 0.00 & 64.30 ± 0.00 & 29.30 ± 0.00 & 32.90 ± 0.00 \\
        \rowcolor{white}
        {} & Head Finetuning (HFT) & 73.90 ± 0.14 &	36.47 ± 0.12 &	40.00 ± 0.00 & 67.27 ± 0.15 & 33.60 ± 0.26 & 36.20 ± 0.10 \\

        {} & Full Finetuning (FT) &  80.47 ± 1.11 & 41.13 ± 0.41 & 44.17 ± 0.39 & 80.73 ± 0.06 & 42.17 ± 0.78 & 44.17 ± 0.25 \\
        
        \cmidrule(lr){2-8}
        \rowcolor{white}
        {} & Visual Prompt (Fixed) & 45.60 ± 0.08 &  21.90 ± 0.08 &  23.77 ± 0.05 & 64.27 ± 0.06 &  29.47 ± 0.06 &  32.83 ± 0.12 \\
        
        \multirow{4}{*}[2em]{\textbf{FLIR - IR}} & Visual Prompt (Random) &  43.27 ± 0.29 & 20.63 ± 0.17 &  22.47 ± 0.12 & 64.03 ± 0.08 & \textbf{29.50 ± 0.00} & 32.77 ± 0.15 \\
        \rowcolor{white}
        {} & Visual Prompt (Padding) & 45.63 ± 1.44 &  22.13 ± 1.08 &  23.87 ± 0.91 & 61.73 ± 0.23 &    27.60 ± 0.17 &  31.20 ± 0.20 \\

        {} & Visual Prompt (WM) & 54.43 ± 0.78 & 26.37 ± 0.54 &  28.67 ± 0.40 & 54.73 ± 0.08 & 23.20 ± 0.10 &  27.20 ± 0.10 \\
        
        \rowcolor{white}
        {} & Visual Prompt (WM$_{v2}$) & 52.43 ± 0.50 &  25.10 ± 0.22 &  27.50 ± 0.22 & 54.80 ± 0.10 &  23.13 ± 0.21 &  27.27 ± 0.06 \\

        \cmidrule(lr){2-8}
        {} & \textbf{ModPrompt (Ours)} & \textbf{69.03 ± 1.06} & \textbf{34.87 ± 0.40} & \textbf{37.33 ± 0.12} & \textbf{65.03 ± 0.05} & 29.23 ± 0.55 & \textbf{32.90 ± 0.26} \\

        \bottomrule
    \end{tabular}
    }
    \caption{Detection performance (APs) for YOLO-World and Grounding DINO for the FLIR-IR dataset. The different visual prompt adaptation techniques are compared with our ModPrompt, and the zero-shot (ZS), head finetuning (HFT), and full finetuning (FT) are also reported, where the full finetuning is the upper bound.}
    \label{tab:main_yolo_world_flir}

\end{table*}

\section{Additional Details of Vision-Language ODs}
\label{sec:additional_details}

For the YOLO-World, we use AdamW optimizer with a learning rate $2e^{-4}$, weight decay $0.05$, and batch size $8$. And for the Grounding-DINO, we use AdamW optimizer with a learning rate $1e^{-4}$, weight decay $1e^{-4}$, and batch size $8$. For the main manuscript, we used YOLO-World Small and Grounding-DINO Tiny. For the experiments with text, we extract the embeddings and optimize them without the text encoder for efficient adaptation of the embedding space. Additionally, we provide results with bigger backbones to further corroborate our findings.

\begin{table*}[!ht]
    \centering
    \resizebox{1.0\textwidth}{!}{%
    \begin{tabular}{lqgggqgg}
        \toprule
        \rowcolor{white}
        {\multirow{2}{*}[-0.7em]{\textbf{Detector}}} & \multirow{2}{*}[-0.7em]{\textbf{Method}}  &  \multicolumn{3}{c}{\textbf{FLIR-IR}} &   \multicolumn{3}{c}{\textbf{NYU$_{v2}$-DEPTH}} \\

        \cmidrule(lr){3-5}
        \cmidrule(lr){6-8}
        \addlinespace[5pt]        
        \rowcolor{white}
        {}  & {} & \multicolumn{1}{c}{\multirow{2}{*}[1em]{\textbf{AP$_{50}$}}} & 
        \multicolumn{1}{c}{\multirow{2}{*}[1em]{\textbf{AP$_{75}$}}} &
        \multicolumn{1}{c}{\multirow{2}{*}[1em]{\textbf{AP}}} & \multicolumn{1}{c}{\multirow{2}{*}[1em]{\textbf{AP$_{50}$}}} & 
        \multicolumn{1}{c}{\multirow{2}{*}[1em]{\textbf{AP$_{75}$}}} &
        \multicolumn{1}{c}{\multirow{2}{*}[1em]{\textbf{AP}}}   \\
        \midrule

        {} &  Zero-Shot (ZS)  & 71.60 ± 0.00 &   37.60 ± 0.00 &  39.30 ± 0.00 &  05.30 ± 0.00 & 03.70 ± 0.00 & 03.50 ± 0.00 \\

        \rowcolor{white}
        {} &  Head Finetuning (HFT) & 82.27 ± 0.21 & 43.53 ± 0.12 &	45.93 ± 0.05 &	24.43 ± 0.17 &	14.63 ± 0.29 &	14.53 ± 0.17 \\

        {} & Full Finetuning (FT) & 84.33 ± 0.45 &	44.00 ± 0.29 &	46.70 ± 0.22 &	54.13 ± 0.05 &	40.43 ± 0.09 &	37.33 ± 0.17 \\

        \cmidrule(lr){2-8}
        \rowcolor{white}
        {} & Visual Prompt (Fixed)  & 71.83 ± 0.09 &	37.70 ± 0.00 &	39.40 ± 0.00 &	05.20 ± 0.00 &	03.60 ± 0.00 &	03.40 ± 0.00 \\\

        \multirow{2}{*}[1em]{\textbf{YOLO-World-Large}} & Visual Prompt (Random)  & 71.40 ± 0.14 &	37.40 ± 0.08 &	39.23 ± 0.05 &	04.67 ± 0.12 &	03.17 ± 0.09 &	02.97 ± 0.09 \\
        
        \rowcolor{white}        
        {} & Visual Prompt (Padding)  & 66.27 ± 0.12 &	33.83 ± 0.05 &	35.97 ± 0.05 &	02.87 ± 0.12 &	01.80 ± 0.08 &	01.80 ± 0.08 \\

        {} & Visual Prompt (WM)  & 71.70 ± 0.14 &	37.53 ± 0.25 &	39.43 ± 0.05 &	15.10 ± 0.45 &	09.83 ± 0.26 &	09.50 ± 0.24 \\

        \rowcolor{white}
        {} & Visual Prompt (WM$_{v2}$)  & 70.90 ± 0.08 &	36.73 ± 0.19 &	38.73 ± 0.05 &	14.53 ± 0.41 &	09.57 ± 0.33 &	09.13 ± 0.31 \\

        \cmidrule(lr){2-8}
        {} & \textbf{ModPrompt (Ours)}  & \textbf{77.13 ± 0.29} & \textbf{41.37 ± 0.69} &	\textbf{43.23 ± 0.24} &	\textbf{39.27 ± 0.53} &	\textbf{28.93 ± 0.17} &	\textbf{26.73 ± 0.09} \\

        \midrule
        \addlinespace[2pt]        
        \midrule

        {} &  Zero-Shot (ZS)  & 64.80 ± 0.00 &   29.10 ± 0.00 &  32.90 ± 0.00 & 08.50 ± 0.00 & 05.70 ± 0.00 & 05.40 ± 0.00 \\

        \rowcolor{white}
        {} &  Head Finetuning (HFT) & 68.40 ± 0.14 & 34.60 ± 0.28 &  36.95 ± 0.21 & 08.50 ± 0.00 &  06.10 ± 0.10 &  05.67 ± 0.06 \\

        {} & Full Finetuning (FT) & 81.97 ± 0.25 &    34.60 ± 0.28 &  45.85 ± 0.07 & 53.93 ± 0.40 &  40.53 ± 0.15 &  37.50 ± 0.26 \\

        \cmidrule(lr){2-8}
        \rowcolor{white}
        {} & Visual Prompt (Fixed)  &  64.87 ± 0.06 &   29.17 ± 0.06 &  33.00 ± 0.00  & 08.53 ± 0.06 &  05.73 ± 0.06 &  05.53 ± 0.06 \\

        \multirow{2}{*}[1em]{\textbf{Grounding DINO-Big}} & Visual Prompt (Random)  & 64.83 ± 0.06 &   29.13 ± 0.06 &  32.93 ± 0.06  & 08.53 ± 0.06 &  05.77 ± 0.15 &  05.47 ± 0.06 \\
        
        \rowcolor{white}        
        {} & Visual Prompt (Padding)  &  62.63 ± 0.06 &   27.27 ± 0.06 &  31.53 ± 0.06  & 07.93 ± 0.06 &  05.13 ± 0.06 &  04.83 ± 0.06 \\

        {} & Visual Prompt (WM)  &  56.77 ± 0.31 &   21.93 ± 0.55 &  27.00 ± 0.36 & 05.57 ± 0.06 &  03.23 ± 0.06 &  03.17 ± 0.06 \\

        \rowcolor{white}
        {} & Visual Prompt (WM$_{v2}$)  &  57.03 ± 0.40 &   22.20 ± 0.40 &  27.20 ± 0.30 & 05.73 ± 0.06 &  03.33 ± 0.06 &  03.33 ± 0.06 \\

        \cmidrule(lr){2-8}
        
        {} & \textbf{ModPrompt (Ours)}  & \textbf{65.73 ± 0.15} &   \textbf{30.07 ± 0.12} &  \textbf{33.47 ± 0.12} & \textbf{25.30 ± 0.53} & \textbf{17.60 ± 0.20} & \textbf{16.57 ± 0.35} \\

        \bottomrule

    \end{tabular}
    }
    \caption{Detection performance (APs) for YOLO-World-Large and Grounding DINO-B on FLIR-IR and NYU$_{v2}$-Depth datasets. Each visual prompt adaptation strategy is compared with our ModPrompt.}
    \label{tab:different_detectors}
\end{table*}

\section{Formal definition of Text-Prompt Tuning}
\label{sec:formal_definition_tpt}

Following our definitions of visual prompts for object detection in the main manuscript, we define the text-prompt tuning using our notation in this section. YOLO-World follows the YOLOv8 loss from~\citet{yolov8_ultralytics}, with a text contrastive head to provide the object-text similarities; for more details about YOLO-World loss check~\citep{cheng2024yolo}. Here, we provide a generic definition, independent of the model. Thus, we define the text-prompt cost function ($\mathcal{C}_{\text{tp}}(\phi)$), with the following Equation: 
\begin{equation} 
\begin{split}
        \mathcal{C}_{\text{tp}}(\phi) =\frac{1}{|\mathcal{D}|} \sum_{(x_{t}, Y) \in \mathcal{D}} \mathcal{L}_{text}(g_\psi(x_{t}+h_{\phi}), Y),
\end{split}
\label{eq:modprompt_embeddingtuning}
\end{equation}
\noindent where $x_{t}$ is the input text, $g_\psi$ is the text-encoder, $h_{\phi}$ is the additional prompt parametrized by $\phi$. In the case of YOLO-World, $\mathcal{L}_{text}$ can be seen as label assignment of~\citet{feng2021tood} to match the predictions with ground-truth annotations, with  Binary Cross Entropy (BCE), and assign each positive prediction with a text index as the classification label.

\section{Additional Results}
\label{sec:results}

\subsection{Main Results on FLIR-IR data:}
\label{sec:results_flir}

In~\tref{tab:main_yolo_world_flir} we compare the performance of our method against the baselines on FLIR-IR dataset. It can be observed that our ModPrompt achieves the highest performance in terms of APs for YOLO-World, and for Grounding DINO the AP$_{50}$ and AP results were the highest, while the AP$_{75}$ is equally good as the random prompt. The FLIR-IR dataset is a more challenging dataset composed of $3$ classes, some small bounding boxes, and missing annotations, which make the problem more difficult when the input image is being changed. We observe that ModPrompt performs better when objects are well-defined in the image and when objects are not too small, otherwise, like all other input-level pixel strategies it struggles, especially on refined bounding-box localization, which can be seen with AP$_{75}$ and AP, whereas in AP$_{50}$ it always shows good results.

\subsection{Results with Different Detection Backbones:}
\label{sec:different_detectors}

In this section, we provide results for the YOLO-World-Large and Grounding DINO-Big models with different visual prompt strategies and our ModPrompt. In~\tref{tab:different_detectors}, we show that ModPrompt is better than all visual prompt methods for FLIR and NYU$_{v2}$.

\subsection{MPDR with FLIR-IR data:}
\label{sec:inference_friendly}

In~\tref{tab:text_residual_study_flir}, we provide additional results for FLIR-IR with our MPDR module. We emphasize that the knowledge preservation strategy improves performance in many cases. However, this dataset is too noisy, which compromises translation methods such as ModPrompt, resulting in degradation of performance in some cases.

\begin{table*}[!ht]
    \centering
    \begin{tabular}{lqggg}
        \toprule
        \rowcolor{white}
        {\multirow{2}{*}[-0.7em]{\textbf{Detector}}} & \multirow{2}{*}[-0.7em]{\textbf{Method}}  &  \multicolumn{3}{c}{\textbf{FLIR-IR}} \\

        \cmidrule(lr){3-5}
        \addlinespace[5pt]        
        \rowcolor{white}
        {}  & {} & \multicolumn{1}{c}{\multirow{2}{*}[1em]{\textbf{AP$_{50}$}}} & 
        \multicolumn{1}{c}{\multirow{2}{*}[1em]{\textbf{AP$_{75}$}}} &
        \multicolumn{1}{c}{\multirow{2}{*}[1em]{\textbf{AP}}} \\
        \midrule

        {} &  \textbf{Fixed} & 63.93 ± 0.26 (+0.63) &  31.90 ± 0.08 (-0.60) &  34.63 ± 0.05 (+0.33) \\

        \cmidrule(lr){2-5}
        \rowcolor{white}
        \multirow{2}{*}[-0.7em]{\textbf{YOLO-World}} & \textbf{Random} & 63.30 ± 0.14 (+0.43) &  31.73 ± 0.09 (-0.44) &  34.27 ± 0.12 (+0.17) \\

        \cmidrule(lr){2-5}
        {} & \textbf{Padding} &  59.23 ± 0.12 (+0.66) &  29.10 ± 0.08 (-0.10) &  31.50 ± 0.08 (+0.17) \\

        \cmidrule(lr){2-5}
        \rowcolor{white}
        {} & \textbf{WeightMap}  &  63.10 ± 0.28 (+0.33) & 31.60 ± 0.29 (-0.10) &  34.20 ± 0.08 (+0.33) \\

        \cmidrule(lr){2-5}
        {} & \textbf{ModPrompt} & \textbf{72.73 ± 0.00 (-1.74)} &  \textbf{37.70 ± 0.00 (-0.60)} &  \textbf{40.37 ± 0.00 (-0.73)} \\

        \midrule
        \addlinespace[2pt]        
        \midrule

        {} & \textbf{Fixed}  & 66.53 ± 0.98 (+2.26) & \textbf{32.83 ± 2.50 (+3.36)} & \textbf{34.83 ± 0.90} (+2.00) \\

        \cmidrule(lr){2-5}
        \rowcolor{white}
        \multirow{2}{*}[-0.7em]{\textbf{Grounding DINO}} & \textbf{Random}  & 66.10 ± 1.08 (+2.07) & 31.40 ± 1.31 (+1.90) & 34.53 ± 0.90 (+1.76) \\
        
        \cmidrule(lr){2-5}
        
        {} & \textbf{Padding}  & 63.33 ± 1.10 (+1.60) & 29.73 ± 1.27 (+2.13) & 32.93 ± 0.90 (+1.73) \\

       \cmidrule(lr){2-5}
       \rowcolor{white}
        {} & \textbf{WeightMap}  &  55.60 ± 0.92 (+0.87) & 24.37 ± 0.65 (+1.17) & 28.30 ± 0.66 (+1.10) \\

        \cmidrule(lr){2-5}
        {} & \textbf{ModPrompt} & \textbf{67.80 ± 0.14 (+2.77)} & 31.10 ± 0.31 (+1.87) & 33.70 ± 0.09 (+0.80) \\

        \bottomrule

    \end{tabular}
    \caption{Detection performance (APs) for YOLO-World and Grounding DINO on FLIR-IR data. Each visual prompt adaptation strategy is compared with the learnable MPDR (results in parenthesis are the gain with the MPDR module), which is responsible for updating the new modality embeddings and not changing the original embedding knowledge.}
    \label{tab:text_residual_study_flir}
\end{table*}

\subsection{Ablation Studies on Visual Prompts:}
\label{sec:ablations_vp}

We evaluate different variations of the visual prompt adaptation methods. Specifically, we compare the performance when different input patch sizes are used; for instance, p$_{s} = 30$ refers to a patch size of $30$ pixels. In this study, we tested various patch sizes for each of the visual prompt methods and reported the performance in~\tref{tab:ablation_tab}. We evaluate modprompt using two different translators with U-Net based backbones, MobileNet (MB)~\citep{howard2017mobilenets} and ResNet (RES)~\citep{he2016deep}. Here, we provide the additional results on FLIR-IR dataset. \newline

\begin{table}[!ht]
    \centering
    \resizebox{1.0\columnwidth}{!}{%
    \begin{tabular}{lcqgg}
        \toprule

        \multirow{2}{*}[-0.7em]{\textbf{Method}} & \multirow{2}{*}[-0.7em]{\textbf{Variation}} &   \multicolumn{3}{c}{\textbf{FLIR - IR}} \\
        \cmidrule(lr){3-5}
        \addlinespace[5pt]        
        {} & {} & \multicolumn{1}{c}{\multirow{2}{*}[1em]{\textbf{AP$_{50}$}}} & 
        \multicolumn{1}{c}{\multirow{2}{*}[1em]{\textbf{AP$_{75}$}}} &
        \multicolumn{1}{c}{\multirow{2}{*}[1em]{\textbf{AP}}}   \\
        \midrule
        \rowcolor{white}
        \multirow{2}{*}[0em]{\textbf{Fixed}}        & 30     & 45.60 ± 0.08  & 21.90 ± 0.08 &  23.77 ± 0.05 \\
        \cmidrule(lr){2-5}
        {}                                 & 300    & 29.30 ± 0.37  & 13.50 ± 0.54 &  15.00 ± 0.37 \\
        \midrule
        \rowcolor{white}
        \multirow{2}{*}[0em]{\textbf{Random}}       & 30     & 43.27 ± 0.29  & 20.63 ± 0.17 &  22.47 ± 0.12 \\
        \cmidrule(lr){2-5}
        {}                                 & 300    & 19.13 ± 0.33  & 09.00 ± 0.42 &  09.80 ± 0.33 \\
        \midrule
        \rowcolor{white}
        \multirow{2}{*}[0em]{\textbf{Padding}}      & 30     & 45.63 ± 1.44  & 22.13 ± 1.08 &  23.87 ± 0.91 \\
        \cmidrule(lr){2-5}
        {}                                 & 200    & 00.53 ± 0.12  & 00.17 ± 0.17 &  00.27 ± 0.09 \\
        
        \midrule
        \rowcolor{white}
        \multirow{2}{*}[0em]{\textbf{ModPrompt}}    & MB     & 66.80 ± 0.29  & \textbf{35.23 ± 0.38} &  36.53 ± 0.12 \\
        \cmidrule(lr){2-5}
        {}                                 & RES    & \textbf{69.03 ± 1.06}  & 34.87 ± 0.40 &  \textbf{37.33 ± 0.12} \\

        \bottomrule
    \end{tabular}
    }
    \caption{Detection performance (APs) for YOLO-World on FLIR-IR data. We compared the main visual prompt strategies \textit{fixed}, \textit{random}, \textit{padding}, and ModPrompt. The variations consist of the number of prompt pixels (p$_{s}=30$, $200$ or $300$) and for ModPrompt, the MobileNet (MB) or ResNet (RES).}
    \label{tab:ablation_tab}
\end{table}

\begin{figure}
\centering
\begin{subfigure}[h]{1.0\columnwidth}

    \includegraphics[width=\textwidth]{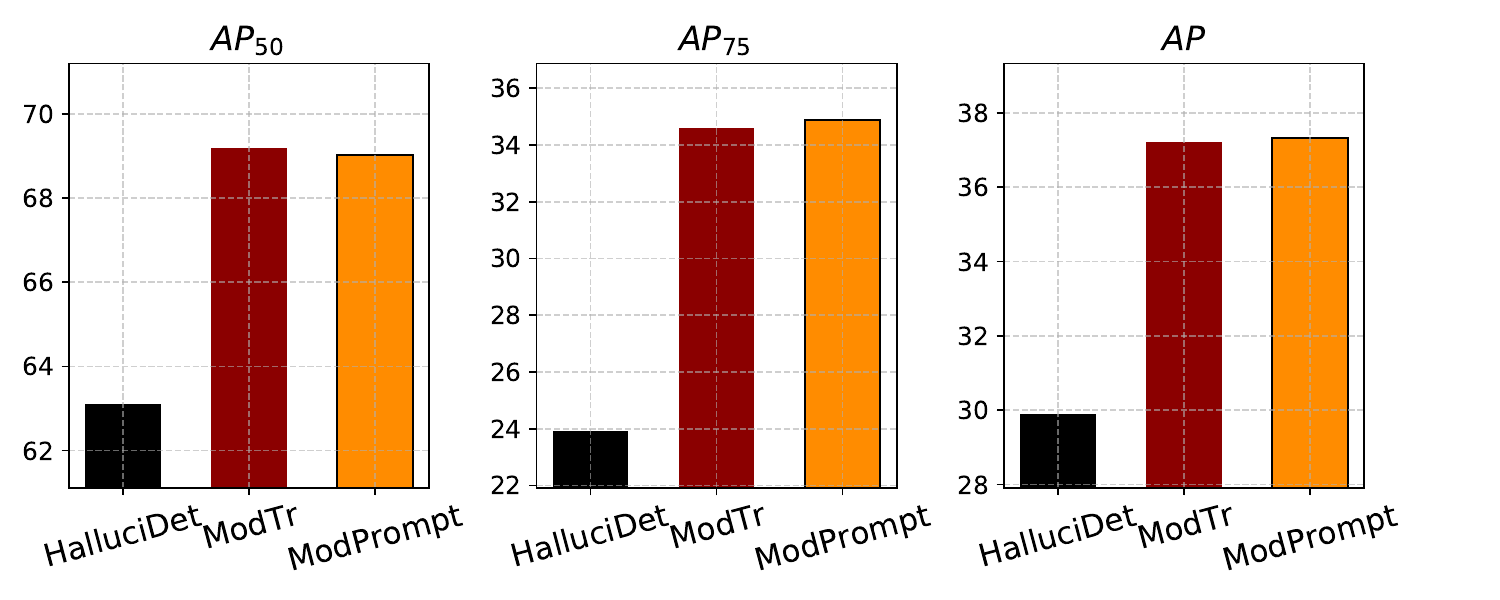}
\end{subfigure}

\caption{Detection performance on FLIR-IR dataset of different modality translators for OD in terms of APs.}
\label{fig:comparative_flir}
\end{figure}

\subsection{Comparison with SOTA Modality Translation OD methods:}
\label{sec:sota}

Our ModPrompt technique is compared with recent state-of-the-art modality translation methods for ODs: HalluciDet~\citep{medeiros2024hallucidet} and ModTr~\citep{medeiros2024modality}. In~\cref{fig:comparative_flir}, we observe that our results are better in all APs for the FLIR dataset.

\subsection{Qualitative Results:}
\label{sec:additional_qualitative_results}

In this section, we provide more visual results for the methods compared, where the performance of each model can be shown by the bounding box predictions. For instance, in~\fref{figure:initial_fig}, we can see that Zero-Shot (ZS) is performing well on the FLIR-IR dataset apart from some false positives. However, the ZS doesn't perform well on the modality that is too different from the pre-training weight, such as depth (NYU$_{v2}$ dataset). For the Visual Prompt (weight map), we have some false positives and missing detections (e.g., the person in the first row for LLVIP and a wrong bounding box). For ModPrompt, we see that we have some good overall detections because the encoder-decoder architecture tends to suppress a little bit of background, as we can see in the first row or second row, but the bounding boxes are not as precise as the right ones in the ZS (see first-row comparison between ZS and ModPrompt). Additionally, in~\fref{fig:qualitative_results}, we provide a batch of $8$ images from the test, we can observe a similar trend for depth and IR modalities as discussed above. Surprisingly, in some cases of the FLIR dataset (challenging dataset with really hard small bounding boxes and some missing labels), our method tends to detect objects that are not labeled in the ground truth (for instance a small person in the first row and first column, behind the two cars on the left, which are not labeled, but our method get its right). This shows the effectiveness of our method and exemplifies the ability of visual language detectors to detect unseen objects.

\begin{figure*}[!ht]
\centering
    \begin{tabular}{c}
  \toprule
    NYU$_{v2}$-Depth \\
    \midrule 

    \makebox[0pt][r]{\makebox[10pt]{\raisebox{20pt}{\rotatebox[origin=c]{90}{\scriptsize GT}}}}
    \includegraphics[width=0.95\textwidth]{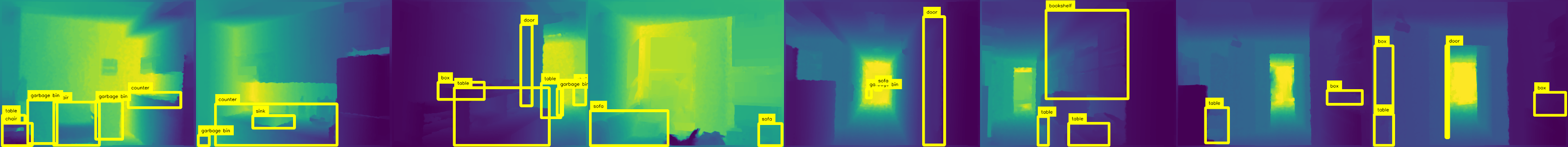} \\
    
    \makebox[0pt][r]{\makebox[10pt]{\raisebox{20pt}{\rotatebox[origin=c]{90}{\scriptsize ZS}}}}
    \includegraphics[width=0.95\textwidth]{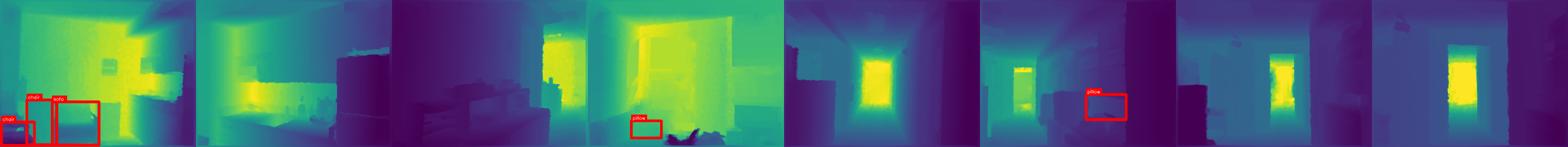} \\
    
    \makebox[0pt][r]{\makebox[10pt]{\raisebox{20pt}{\rotatebox[origin=c]{90}{\scriptsize VP}}}}
    \includegraphics[width=0.95\textwidth]{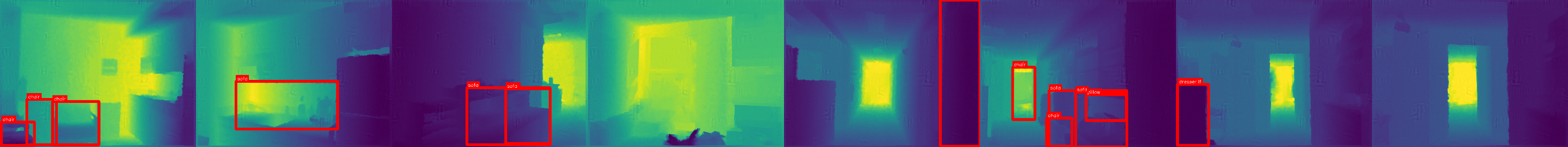} \\
    
    \makebox[0pt][r]{\makebox[10pt]{\raisebox{20pt}{\rotatebox[origin=c]{90}{\scriptsize MP}}}}
    \includegraphics[width=0.95\textwidth]{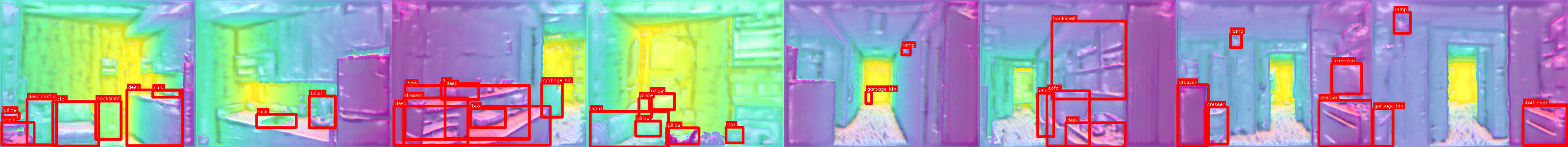} \\ 

    \midrule
    FLIR-IR \\
    \midrule

    \makebox[0pt][r]{\makebox[10pt]{\raisebox{20pt}{\rotatebox[origin=c]{90}{\scriptsize GT}}}}
    \includegraphics[width=0.95\textwidth]{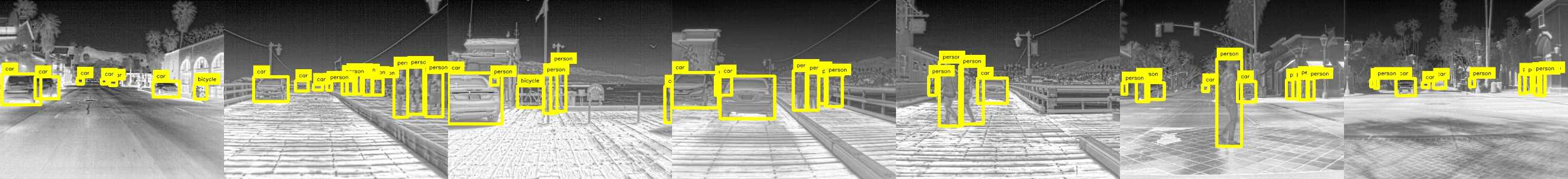} \\
    
    \makebox[0pt][r]{\makebox[10pt]{\raisebox{20pt}{\rotatebox[origin=c]{90}{\scriptsize ZS}}}}
    \includegraphics[width=0.95\textwidth]{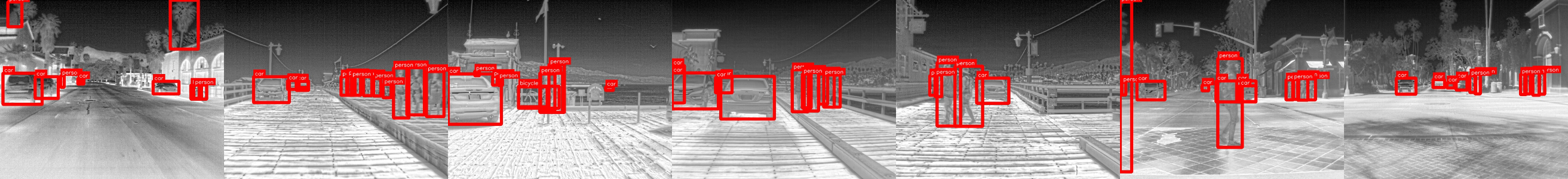} \\
    
    \makebox[0pt][r]{\makebox[10pt]{\raisebox{20pt}{\rotatebox[origin=c]{90}{\scriptsize VP}}}}
    \includegraphics[width=0.95\textwidth]{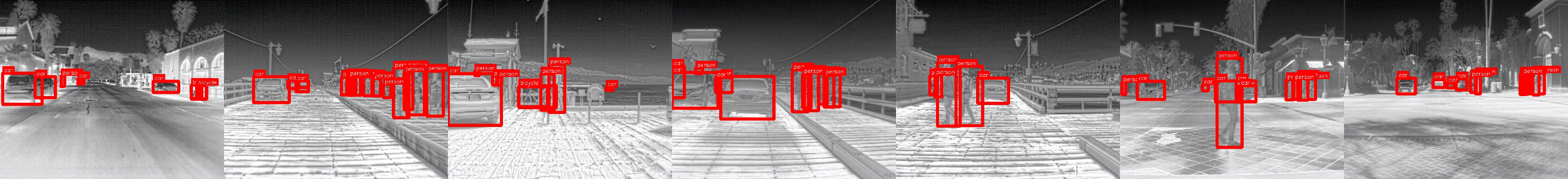} \\
    
    \makebox[0pt][r]{\makebox[10pt]{\raisebox{20pt}{\rotatebox[origin=c]{90}{\scriptsize MP}}}}
    \includegraphics[width=0.95\textwidth]{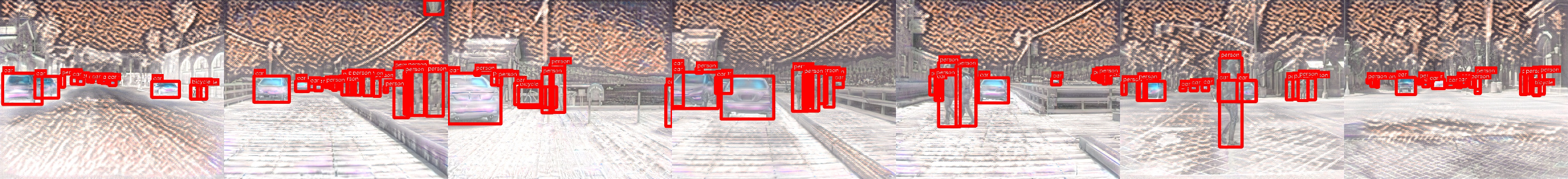} \\

    \bottomrule

    \end{tabular}
\caption{\textbf{Detections of different approaches across modalities for YOLO-World}: NYU$_{v2}$ (depth) and FLIR (infrared). Each row corresponds to a different approach: \textbf{GT (Ground Truth):} Shows in yellow the ground-truth bounding boxes for objects. \textbf{ZS (Zero-Shot):} Displays detections (in red) from a zero-shot model YOLO-World-s.  \textbf{VP (Visual Prompt):} Illustrates detections with weight map visual prompt added to the image. \textbf{MP (ModPrompt):} Detections from our proposed model.}

\label{fig:qualitative_results}

\end{figure*}

\section{Limitations and future works}
\label{sec:limitations_futureworks}

\textbf{Limitations:} In this section, we argue that the work still has some limitations, which we believe can be further improved in subsequent works. For instance, the adaptation strategies still require target labels, which could be explored under other tasks, such as unsupervised or semi-supervised approaches. Additionally, we argue that our method is still not perfect for small objects and it incorporates some noise which can be further minimized by other loss constraints if we had access to additional source data (which we didn't during training). Some of the limitations were already discussed in the qualitative results, which can be summarized as difficult under small objects and duplications of bounding box predictions. \newline

\noindent\textbf{Future works:} Future works could improve the conditioning on both text and vision, and exploit more label-efficient adaptation strategies such as test-time adaptation or few-shot learning. \newline

\clearpage

{
    \small
    \bibliographystyle{ieeenat_fullname}
    \bibliography{main}
}

\end{document}